\UseRawInputEncoding

\documentclass[lettersize,journal]{IEEEtran}
\usepackage{amsmath,amsfonts}
\usepackage{algorithmic}
\usepackage{algorithm}
\usepackage{array}
\usepackage[caption=false,font=normalsize,labelfont=sf,textfont=sf]{subfig}
\usepackage{textcomp}
\usepackage{stfloats}
\usepackage{url}
\usepackage{verbatim}
\usepackage{graphicx}
\usepackage{booktabs}
\usepackage{cite}
 \usepackage{mathrsfs}
\usepackage{diagbox}
\usepackage{threeparttable}
\usepackage{multirow}
\usepackage[T1]{fontenc}
\usepackage{amssymb}
\usepackage{fancyhdr}
\def\BibTeX{{\rm B\kern-.05em{\sc i\kern-.025em b}\kern-.08em
    T\kern-.1667em\lower.7ex\hbox{E}\kern-.125emX}}
\usepackage{balance}
\begin{document}
\title{A Multi-Task Deep Learning Approach for Sensor-based Human Activity Recognition and Segmentation }
\author{Furong Duan, Tao Zhu,~\IEEEmembership{Member,~IEEE},  Jinqiang Wang,
	Liming Chen,~\IEEEmembership{Senior Member,~IEEE,} Huansheng Ning,~\IEEEmembership{Senior Member,~IEEE,} and Yaping Wan
\thanks{
This work is partly supported by the National Natural Science Foundation of
China (62006110, 62071213).The Research Foundation of Education Bureau
of Hunan Province (21C0311, 21B0424). Hengyang Science and Technology
Major Project: 202250015428. (Corresponding author: Tao Zhu.)

Furong Duan, Tao Zhu, Jinqiang Wang, and Yaping Wan were with the Department of Computer Science, University of South China, 421001 China (email: frduan@stu.usc.edu.cn, tzhu@usc.edu.cn, jqwang@stu.usc.edu.cn, ypwan@aliyun.com).

Liming Chen was with the School of Computing and Mathematics, University of Ulster, Belfast, BT37 0QB, U.K. Huansheng Ning was Department
of Computer \& Communication Engineering, University of Science and
Technology Beijing, 100083 China (email: l.chen@ulster.ac.uk, ninghuansheng@ustb.edu.cn).
}}
\maketitle
\thispagestyle{fancy}
\cfoot{\small{This work has been submitted to the IEEE for possible publication. Copyright may be transferred without notice, after which this version may no longer be accessible.}}
\begin{abstract}
Sensor-based human activity segmentation and recognition are two important and challenging problems in many real-world applications and they have drawn increasing attention from the deep learning community in recent years. 
Most of the existing deep learning works were designed based on pre-segmented sensor streams and they have treated activity segmentation and recognition as two separate tasks. In practice, performing data stream segmentation is very challenging.  We believe that both activity segmentation and recognition may convey unique information which can complement each other to improve the performance of the two tasks. In this paper, we firstly proposes a new multitask deep neural network to solve the two tasks simultaneously. The proposed neural network adopts selective convolution and features multiscale windows to segment activities of long or short time durations. First, multiple windows of different scales are generated to center on each unit of the feature sequence. Then, the model is trained to predict, for each window, the activity class and the offset to the true activity boundaries. Finally, overlapping windows are filtered out by non-maximum suppression, and adjacent windows of the same activity are concatenated to complete the segmentation task.
Extensive experiments were conducted on eight popular benchmarking datasets, and the results show that our proposed method outperforms the state-of-the-art methods both for activity recognition and segmentation.

\end{abstract}


\section{INTRODUCTION}
\IEEEPARstart{S}{ensor}-based activity segmentation and recognition algorithms have been widely employed in real-life applications, such as ambient-assisted living\cite{chen2012sensor,chen2011knowledge, cook2009ambient}, human-computer interaction\cite{lara2012survey, krishnan2014activity, wang2017human} and healthcare \cite{kim2009human,rashidi2010discovering, acampora2013survey}. An activity segmentation algorithm concentrates on identifying the starting and ending positions of the activity, and an activity recognition algorithm aims to recognize the activities based on the data streams collected from the various sensors.

In the field of activity segmentation, several approaches use a fixed window to segment the activity, and the selection of the segmented window length is based on experience. A large window contains more information about the data stream but may fail to identify activity transitions if there are different activities within the window. In contrast, a small window can quickly capture activity transitions but may lead to misclassifications because it does not contain enough information.
To tackle this dilemma, some researchers try to adaptively determine the window length based on the probability of the signal belongs to a particular activity \cite{noor2017adaptive}. Some researchers proposed a time window based segmentation model and dynamically manipulated the model parameters to shrink or expand the time window’s length  \cite{okeyo2014dynamic}.

In the field of activity recognition, many traditional machine learning algorithms, such as SVM, KNN and random forest, apply handcrafted statistical features to represent the characteristics of the data stream, and have achieved promising performances. However, extracting the effective features depends on domainnon expertise.
Recently, deep learning techniques, which can automatically learn rich features, have been widely applied in sensor-based activity recognition tasks\cite{wang2021sensor}\cite{wang2022negative}. CNNs\cite{ronao2016human}, LSTM\cite{hochreiter1997long}, DeepConvLSTM\cite{ordonez2016deep}, and DeepConvLSTM-Attention\cite{zhang2020sensors} models have been demonstrated to improve the accuracy of activity recognition.


However, most of the earlier works are designed based on pre-segmented sensor streams and treat activity segmentation and recognition as two separate tasks. In real life, 
Since activities do not have a predefined sequence and duration, a pre-segmented offline segmentation technique is performed by examining all the datasets and therefore is not suitable for real-time applications.

Furthermore, if data segmentation is considered a preprocessing step, errors in data segmentation may propagate to the later steps\cite{qian2021weakly}. We believe that both activity segmentation and recognition may convey unique information that can complement each other to improve the performance of activity segmentation and activity recognition.  In this paper, we propose a novel multi-task framewok for  simultaneously sensor based activity segmentation and recognition. The framewok is motivated by the SSD\cite{liu2016ssd} method that exploited in the computer vision field. First, we generate multiple windows of different scales centered on each unit of the activity feature sequence. Second, we formulate a deep neural network to simultaneously predict the activity class contained in each window and the offset of the window from the truth activity boundary. Next, considering the same surrounding activity, many predicted windows with high overlap may be output. we use the non-maximum suppression algorithm \cite{girshick2014rich} to keep the appropriate window. Finally, we combine activity segmentation with activity recognition to jointly update the learnable parameters of the different components. The main contributions of this paper are summarized as follows:
\begin{itemize}
	\item We propose a novel multi-task framewok, denoted by MTHARS, which can effectively combine activity segmentation and recognition to improve the performance of two tasks.
	\item Taking the dynamics of activity length, we propose a multiple scale windows concatenation method to segment activities of various length.
	\item We use eight popular benchmarking activity datasets for the activity segmentation and recognition experiments. The experimental results show that MTHARS performs better than the state-of-the-art methods. An ablation study is to analyze the effects of several key factors.
\end{itemize}

The rest of the paper is organized as follows. 	Section \uppercase\expandafter{\romannumeral2} describes related works. Section \uppercase\expandafter{\romannumeral3} the proposed method. Section \uppercase\expandafter{\romannumeral4} presents the eight popular benchmarking datasets and the experimental setup, specifically, the experimental results for the activity segmentation and recognition and the ablation study. Section \uppercase\expandafter{\romannumeral5} summarizes our work and discusses future work.

\section{RELATED WORK}
In this section, related works are presented from two aspects, activity segmentation and activity recognition.

\subsection{Activity Segmentation}
Change point detection(CPD) based methods are used to detect the point in time when the behavior of a time series abruptly changes \cite{aminikhanghahi2018real}.
When a supervised approach is employed for change point detection, machine learning algorithms
can be trained as binary or multi-class classifiers\cite{aminikhanghahi2017survey}.
 \cite{guedon2013exploring} proposed computing several most likely fragment candidates based on the reversibility of the time series data. 

The sliding window-based sensor data stream segmentation method is a common method for segmenting sensor data streams. A fixed window length is used to proportion the data stream into the same size segments for input into the model for activity recognition. The window length determines the length of each data segment, which directly affects the quality of the feature extractions and the accuracy of the classifications performed by the model \cite{chen2021deep}. Many strategies for segmenting sensor data based on sliding windows have been proposed. 

Wang et al.\cite{wang2012hierarchical} divided the entire sequence of
sensor events into equal size time intervals. Banos et al.\cite{banos2014window} experimented with various window sizes ranging from 0.25 s to 7 s in steps of 0.25 s on different activities. However, since the various activities are completed in different time periods, defining the optimal window size is challenging.

Aminikhanghahi and Cook\cite{aminikhanghahi2018real} proposed that the sliding window length should be determined by whether there is a sufficient difference between the adjacent windows. Okeyo and Liming Chen\cite{okeyo2014dynamic} presented a time window based segmentation model and dynamically manipulated the model parameters to shrink or expand the time window’s length. WanJie et al. \cite{wan2015dynamic} adopted a dynamic segmentation approach that identifies simple actions using both the sensor and time correlations. Noor and Mohd Halim Mohd \cite{noor2017adaptive} recommended that the window be automatically resized based on the activity signal distribution.
Hong Li et al. \cite{li2018specialized} recommended starting with a fixed sliding window and then adjusting the window length based on various activities.
Santos et al.\cite{santos2015trajectory} proposed a dynamically adaptive sliding window method, whose window length and overlap are continuously adjusted based on entropy feedback. Congcong et al.\cite{ma2020adaptive} proposed to detect the signal difference between time windows using multivariate gaussian distribution.

JuanYe et al. \cite{ye2015kcar} proposed a new knowledge-driven concurrent activity recognition(KCAR) approach. In KCAR, the basic semantics of sensor events and the semantic similarity are investigated by dividing a continuous sensor sequence into different segments, each corresponding to an ongoing activity.
Triboan and Liming Chen \cite{triboan2017semantic}  applied a semantic-based approach which use ontologies to perform terminology box and assertion box reasoning, along with logical rules to infer whether the incoming sensor event is associated with a given sequences of the activity. Triboan and Liming Chen \cite{triboan2019semantics} proposed a semantics-based 
method to perform  sensor data segmentation by combining generic knowledge and resident-specific preferences  to facilitate the segmentation process. 

\subsection{Activity Recognition}
In the field of activity recognition, CNNs consist of a convolutional, pooling layer and a fully connected layer that can learn unique representations.
Ronao and Cho\cite{ronao2016human} proposed a convnet composed of alter-nating convolution and pooling layers. The basic features were extracted at the lower levels, and the more complicated features were extracted at the higher levels.
Bianchi et al.\cite{bianchi2019iot} adopted a CNN model that consists of four convolutional layers and a fully connected layer for recognizing activities.
 Manuel et al.\cite{gil2020improving} proposed a new CNNs structure considering sensor-independent  handling. Pham et al.\cite{pham2020senscapsnet} designed a senscapsnet approach that is suitable for spatial-temporal data. Cruciani et al. \cite{cruciani2020feature} proposed using a CNN to recognize activity based on sensors and audio for better extraction of the temporal features in activity data. Huang et al.\cite{huang2021shallow} presented a shallower CNN implementation of the information interaction between channels in HAR scenarios, in which graph convolution is employed to implement the information interaction between the channels.

To better learn the temporal and spatial aspects of the activity data, Ordóñez et al. \cite{ordonez2016deep} suggests the use of a CNN mixed with LSTM layers. Cheng et al.\cite{xu2019innohar} proposed a InnoHAR model which combine inception neural network
and RNN. Xia et al.\cite{xia2020lstm} presented that two layers of an LSTM were utilized to extract the temporal data features, followed by two levels of convolution and a combination of pooling layers and a BatchNorm between the convolution layers to extract the spatial features. Tong et al.\cite{tong2022novel} proposed using the Bi-GRU-Inception model to learn the temporal features using a GRU network and Inception to analyze the spatial features between multiple channels.


 In recent years, attention mechanisms have been investigated for deep HAR models. Y. Tang et al.\cite{tang2022triple} offers a ternary attention mechanism that includes three separate attention branches for the channel, time and sensor modalities. Zanobya et al.\cite{Khan2021} posits that generating multiple heads with attention is more advantageous for improving convolutional feature representations, and each head in the multihead model is made up of two convolutional blocks. Existing models that use CNNs for activity recognition typically have the same receptive field size in the neurons of the same layer, limiting the possibility of capturing more features of different sizes in the same stage\cite{gao2021deep}. Using the varying size of the receptive fields on top of the feature maps at different scales was proposed early in computer vision to capture more valuable information. As a result, \cite{gao2021deep} presented SK convolution, which uses the attention approach to learn multiscale features by modifying the size of the perceptual field automatically to learn features of different scales.
\begin{figure*}[htb]
	\centering
	\includegraphics[width=\textwidth]{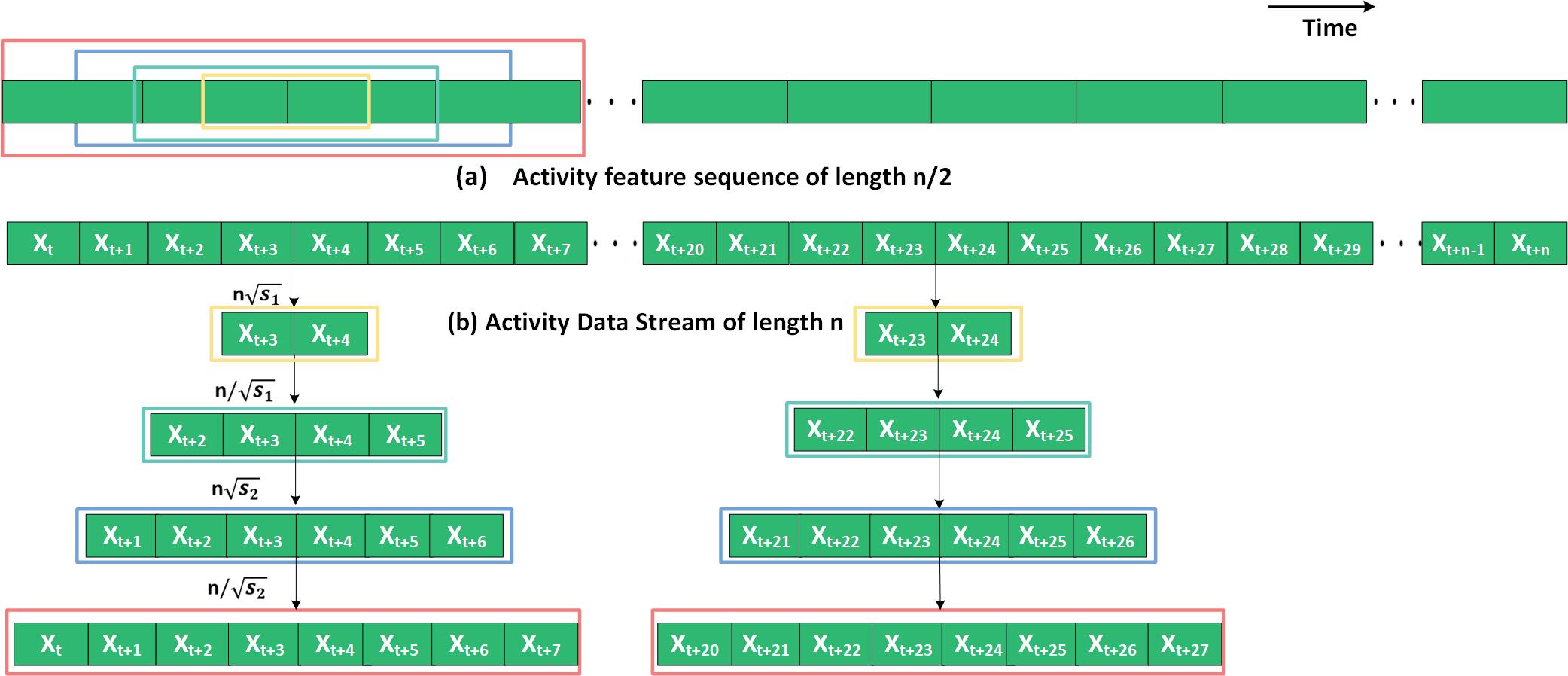}
	
	\caption{(a) We generate four windows of different scales on the feature sequence, with each unit as the center of the window. (b) Four windows of different scales are generated centered on each data point. (The four colors represent the four different lengths of windows.)}
	\label{matching strategy}
\end{figure*}
\section{MTHARS METHODOLOGY}
In this section, we provide details of  the problem definition and proposed MTHARS method inspired by SSD\cite{liu2016ssd} that exploited in the computer vision field. There are four main components: multiscale window generation and matching, non-maximum suppression, model architecture, and model training. 
\subsection{Problem definition}
Given a stream of activity data \textbf{X} = \{$\mathbf{x_i}$\}$_{i=1}^N$, where $\mathbf{x_i}$ denotes a vector of signals collected at timestamp $\textit{t}$. $\mathbf{a}$ = \{a$_k$\}$_{k=1}^K$ represents the \textit{K} activity labels in the stream of activity data, where a$_k$ $\in$ \{\textit{}, \textit{A$_2$
}, \textit{A$_3$
} $\cdots$ \textit{A$_L$}\}. The duration of each a$_k$ is different. Our goal is to segment the activity data into \textit{K} nonoverlapping segments \{\textbf{X$_k$}\}$_{k=1}^K$, each \textbf{X$_k$} $\subseteq$ \textbf{X}, and $\bigcup$$_{k=1}^{k=K}$ \textbf{X$_k$} = \textbf{X}. each \textbf{X$_k$} is assigned an activity a$_k$. The start and end of \textbf{X$_k$} represent the activity boundary of the corresponding activity. The activity boundary can be converted to the form of a center coordinate and length (\textit{t$^x$}, \textit{t$^l$})

\subsection{Multiscale windows}
{\bf{Representation of the window}}: Unlike the conventional sliding window used in activity recognition, we leverage  computer vision's anchor concept. A window is denoted by (\textit{x}, \textit{l}), where \textit{x} is the data point denoting the center position of a window, and \textit{l} represents the length of the window.

{\bf{Generation of windows}}: Assume that the provided activity data length is \textit{n}. With a scale of \textit{s} $\in$  (0, 1], We generate windows of various lengths centered on each data point of the activity data. \textit{n}$\sqrt{s}$ and \textit{n}/$\sqrt{s}$ are the generated window lengths. We can use a variety of scales to build several windows of various lengths. \textit{s} takes the values of \textit{s}$_1$, \textit{s}$_2$
$\cdots$ \textit{s}$_m$. When using these scales to generate windows of different lengths centered at each data point, this will generate \textit{n} $\times$ m $\times$ 2 windows. In this way, we generate multiple windows of different lengths to cover all the truth activity with various lengths as much as possible. As seen in Fig. \ref{matching strategy}, we generated four windows of different lengths centered on each data point.

As mentioned above, we generate multiple windows centered on each data point of the input data stream.
However, by directly generating windows at the center of each data point, we will end up with many windows to calculate.
Consider a data stream of length 600 if we generate 4 windows of different lengths centered on each data point. We need to predict 2400 windows on the data stream. To simplify the computational complexity, we use the receptive field in convolutional operations for guidance. Following its insight, by defining the length of the feature sequence generated from the backone network, we are able to determine the center of a uniformly sampled window on any data stream. Therefore, we generate windows centered on each unit of the feature sequence(As seen in Fig. \ref{matching strategy}) and divide the center of the window by the length of the feature sequence. Therefore, the value of \textit{x} indicates the relative position of the window in the feature sequence. Since the centers of the window are distributed over all the units of the feature sequence, these centers are uniformly distributed over any input data stream based on their relative spatial locations.

{\bf{IOU}}: We employ the Jaccard index to measure the similarity between the generated window and the truth activity bounding. The Jaccard index is defined as the ratio of the intersection length of the window and the true activity bounding box to their merge length. For windows and truth activity boundaries, we denote their Jaccard values as an intersection over union (IOU). The range of an IOU is a number between 0 and 1 that indicates whether there is any overlap. A value of 0 indicates that there is no overlap, while a value of 1 indicates that they are equal. For example, given windows \textit{W$_{1}$}(5, 10), \textit{W$_{2}$}(30, 50), and truth activity bounding box \textit{T$_{1}$}(40, 60), the IOU of \textit{W$_{1}$} and \textit{T$_{1}$} is 0, and the IOU of \textit{W$_{2}$} and \textit{T$_{1}$} is 0.69.

{\bf{Multiscale window labeling and matching}}: During the training process, each generated window is treated as a training sample. To train the model, we must label each window with an activity class and a bounding offset, where the former refers to the activity associated with the window and the latter refers to the offset of the truth activity bounding box in relation to the window. We generate multiple windows of various lengths for the feature sequence. Next, we predict the activity class and boundary offset of all the windows, then we adjust their length according to the predicted offsets to better obtain the predicted window. Finally, we only output the predicted activity windows that satisfy the criteria.
\textbf{\begin{figure*}[htb]
		\centering
		\includegraphics[width=\textwidth]{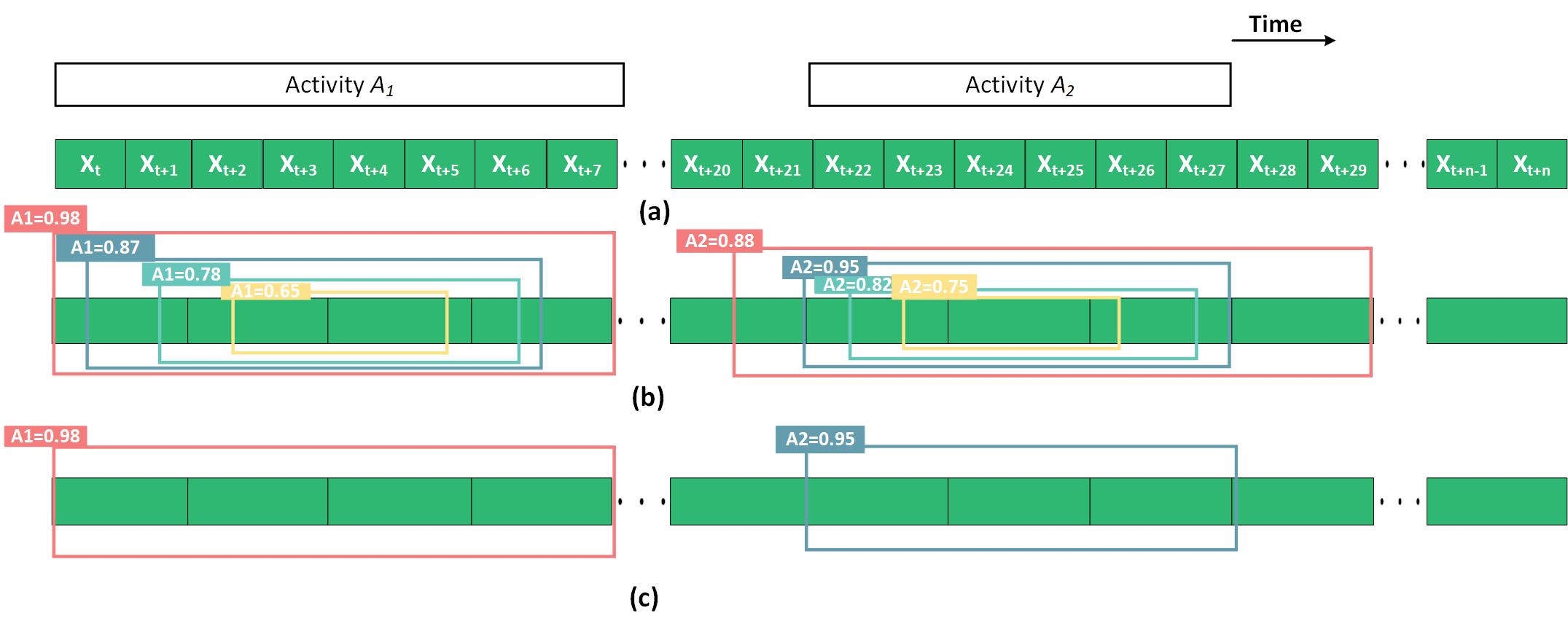}
		\caption{(a) Activity data streams with truth activity boundaries. (b) Windows of different lengths with class probabilities. For example, \textit{A$_{1}$} = 0.98 represents a class probability of 0.98 for activity \textit{A$_{1}$} in this window. (c) Final maintained activity windows by the non-maximum suppression algorithm. }
		\label{non-maximum suppression}
\end{figure*}}
We labeled the training set with the true activity boundary location and the class. To mark any generated window, we refer to the labeled position and class of its assigned truth activity bounding box closest to the window. We apply the following method to assign the closest truth activity bounding box to the window.

Assume the generated windows are \textit{W$_{1}$}, \textit{W$_{2}$} $\cdots$, \textit{W$_{na}$}, and the truth activity bounding boxes are \textit{T$_{1}$},
\textit{T$_{2}$} $\cdots$, \textit{T$_{nb}$}, where \textit{na} $\textgreater$ \textit{nb}. First, we create a matrix \textbf{M} $\in$ $\mathbb{R}$$^{(n_{a}
\times n_{b})}$, in which the element \textit{m$_{ij}$} in the \textit{i$^{th}$} row and \textit{j$^{th}$} column is the IOU of the window \textit{W$_{i}$} and the truth activity bounding boxes \textit{T$_{j}$}. The method's steps are as follows:
\begin{itemize}
	\item Finding the largest element in matrix \textbf{M}, we set its row index and column index to \textit{i$_{1}$} and \textit{j$_{1}$}, respectively, and then assign the truth activity bounding box \textit{T$_{j_{1}}$} to the generated window \textit{W$_{i_{1}}$}, discarding all the elements in the \textit{i$_{1}$} rows and \textit{j$_{1}$} columns of the matrix \textbf{M} after the first assignment.
	\item   Find the largest value of the remaining elements in matrix M. \textit{i$_{2}$}, \textit{j$_{2}$} are the row and column indices, respectively. We assign the generated window \textit{W$_{i_{2}}$} to the truth activity bounding box \textit{T$_{j_{2}}$} and discard the \textit{i$_{2}$} row and all the elements in the \textit{j$_{2}$} column.
	\item Next, we repeat the previous steps until all of the entries in column \textit{nb} of the matrix \textbf{M} have been discarded. At that time, all of the true activity bounding boxes have been allocated a generated window.
	\item Finally, we iterate the remaining \textit{na-nb} windows one by one. If we identify the true activity bounding box \textit{T$_{j}$} with the largest ratio to the window IOU in row \textit{i} of the matrix \textbf{M} and the ratio is greater than the threshold value, we assign the true activity bounding box \textit{T$_{j}$} to window \textit{W$_{i}$}. 
\end{itemize}
For better comprehension of the process, we provide a concrete example to illustrate the process of assigning the generated window to the truth activity bounding box. Considering that the greatest IOU ratio in the matrix \textbf{M}$ \in$ $\mathbb{R}$ $^{(5\times3)}$ is \textit{m$_{51}$}, the truth activity bounding box \textit{T$_{1}$} is assigned to the window \textit{W$_{5}$}. Then, we discard all the elements in Row 5 and Column 1 of the matrix \textbf{M}. 
\[
\textbf{M} = \begin{bmatrix}
	0.55 & 0.82 &  \textbf{0.96}	\\
		0.69 & \textbf{0.95} & 0.78	\\
			0.32 & 0.48 & 0.88	\\
				0.75 & 0.67 & 0.45	\\
					\textbf{0.98} & 0.67 & 0.88	\\
\end{bmatrix}_{5 \times 3}	
\]
Then, we find the maximum \textit{m$_{13}$} in the remaining region of the matrix, assign the truth activity bounding box \textit{T$_{3}$} to the window \textit{W$_{1}$}, and discard all the elements in Row 1 and Column 3 of the matrix. Next, we choose the largest \textit{m$_{22}$} in the remaining part of the matrix to allocate the truth activity bounding box \textit{T$_{2}$} to the window \textit{W$_{2}$} and discard all the components in Row 2 and Column 2. After that, we just need to iterate through the remaining unassigned windows \textit{W$_{3}$}, \textit{W$_{4}$} and assign them a truth activity bounding box based on whether their IOU values with the truth bounding box are greater than the threshold. 

5. We assign the truth activity bounding box to the generated windows. Then, we proceed to annotate the activity class and offset with the following approach. If a window is assigned to a truth activity bounding box, the class of the window will be labeled as that of a truth activity bounding box. The following method is used to compute the true offset.

\begin{figure*}[htb]
	\centering
	\includegraphics[width=\textwidth]{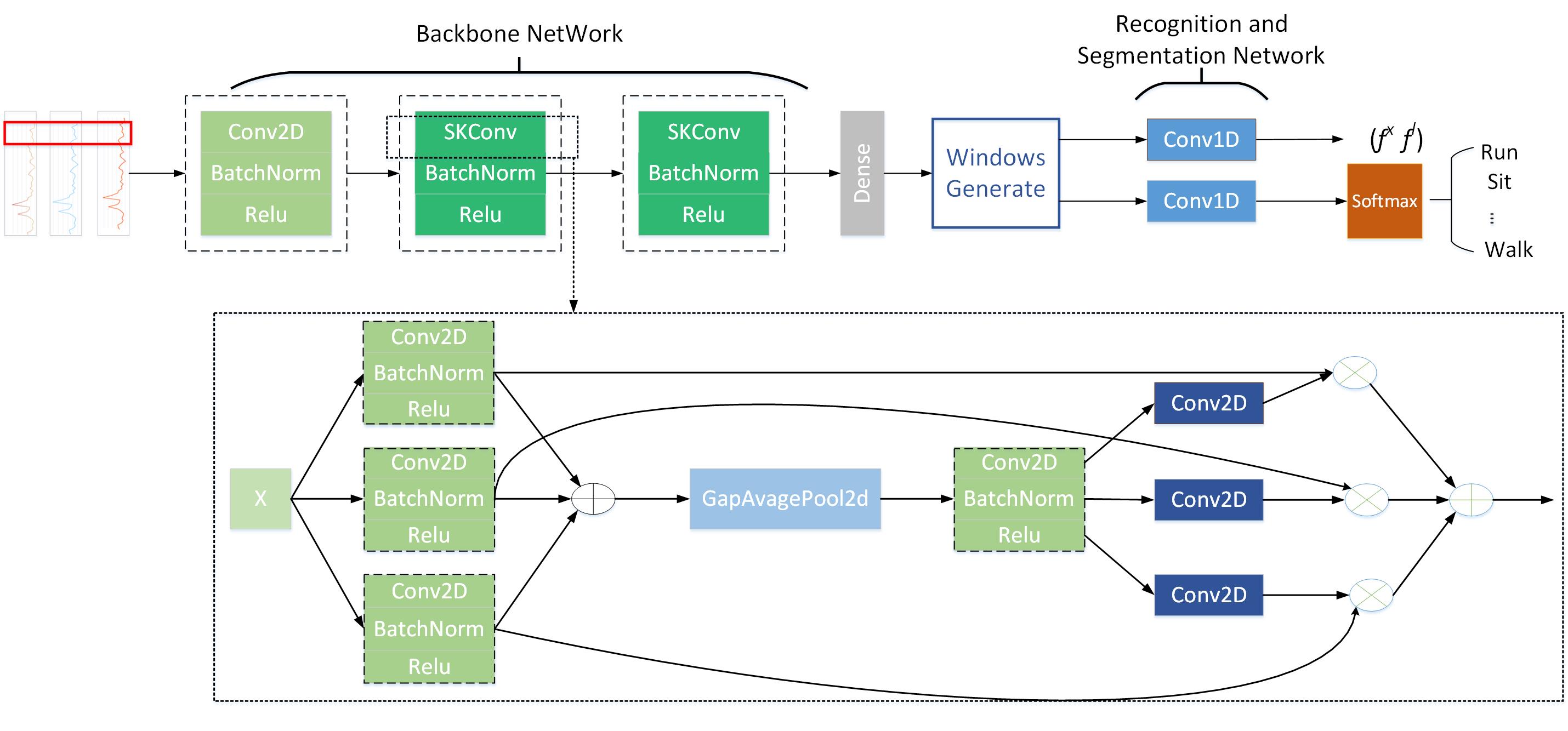}
	\caption{Overview of the Proposed MTHARS Network Architecture}
	\label{segmentmodel}
\end{figure*}

Given a truth activity bounding box \textit{T}= (\textit{$t^{x}$}, \textit{$t^{l}$}) and a window \textit{W} = (\textit{$w^{x}$}, \textit{$w^{l}$}). 
We define the offset as (\textit{f$^{x}$}, \textit{f$^{l}$}), where \textit{f$^{x}$}, \textit{f$^{l}$} represents the center offset and the length offset, respectively, calculated as follows:
	\begin{equation} 
	f^{x} = \frac{t^{x} - w^{x}}{w^{l} }\label{center offset cx}
	\end{equation}
	\begin{equation} 
		f^{l} = \log \frac{t^{l}}{w^{l} }\label{center offset l}
	\end{equation}
{\bf{Predicted activity boundary}}: In the predicting process, we generate multiple windows for each activity and predict the activity class and offsets for each window. The offsets (\textit{f$^{x}$}, \textit{f$^{l}$}) are obtained by the Recognition and Segmentation
NetWork (as in Fig \ref{segmentmodel}). We employ the predicted offsets to calculate the predicted activity boundary (\textit{$\hat{t}$$^{x}$}, \textit{$\hat{t}$${^l}$}) in the following way, where \textit{$\hat{t}$$^{x}$}, \textit{$\hat{t}$${^l}$} denote the activity center and length, respectively.
\begin{equation} 
	\hat{t}^{x} = f^{x} w^l + w^x \label{predictionActivityCx}
\end{equation}
\begin{equation}
	\hat{t}^{l} = w^l\exp(f^{l} )\label{predictionActivityL}
\end{equation}
For example, provided the activity data stream \textbf{X} = \{$\mathbf{x_i}$\}$_{i=1}^N$, we predict that the boundaries of activities \textit{A$_{1}$} and \textit{A$_{2}$} are (100, 200) and (152, 304), respectively, and we can also represent the activity boundary by transforming A and B into the form of the starting position and the ending position, such as (1, 201) and (202, 506). Therefore, \{$\mathbf{x_i}$\}$_{i=1}^{201}$ \{$\mathbf{x_i}$\}$_{i=202}^{506}$ are the activities \textit{A$_{1}$} and \textit{A$_{2}$}, respectively.
	
{\bf{Non-maximum Suppression (NMS):}}
When there are so many windows present, for the same surrounding activity, many predicted windows with high overlap may be output. To simplify the output, we can merge similar windows belonging to the same activity by using a non-maximum suppression algorithm.

The following is the non-maximum suppression algorithm procedure. For a predicted window \textit{W}, the Recognition and Segmentation NetWork predicts a probability \textit{c} for each activity class. The maximum probability \textit{c}  is used to denote the activity class probability corresponding to the window \textit{W} (as shown in Fig.\ref{non-maximum suppression}). Within the same activity, all the predicted windows are sorted by class probability \textit{c} in descending order and organized in the form of a list \textit{L}. Then, we process this sorted list \textit{L} as follows.
\begin{itemize}
	\item Select the window \textit{W$_{1}$} with the highest class probability \textit{c} as the base window and remove the rest of the windows with \textit{W$_{1}$} IOU values exceeding the threshold. Thus, the window \textit{W$_{1}$} with the highest \textit{c} is retained, and the windows with high similarity to \textit{W$_{1}$} are discarded. In other words, the window with the non-highest class probability is suppressed.
	\item The window \textit{W$_{2}$} with the second highest class probability \textit{c} is selected as another base window, and the remaining window with \textit{W$_{2}$}, where the IOU value exceeds the threshold is deleted.
	\item  Repeat the above steps until all windows are selected as base windows or discarded. In this way, the IOU values of any two predicted windows are below the threshold.
\end{itemize}

\begin{table*}[htbp]
	\renewcommand\arraystretch{2.0}
	\centering 
	\caption{Simple Parameter Description of The MTHARS}\label{network description}
	\resizebox{\textwidth}{!}{
		\begin{threeparttable}
		\begin{tabular}{|c|c|c|c|}
		\hline
		\bf{Layers}&\multicolumn{3}{c|}{\bf{Parameters}}\\
		\cline{1-4}
		\bf{Layer1}&\multicolumn{3}{c|}{Conv2D(5$\times$1 / 3$\times$1 / 1$\times$0 )}\\
		\cline{1-4}
		\multirow{3}{*}{\bf{SKConv}}&Conv2D(3$\times$1 / 1$\times$1 / 1$\times$1 / 1$\times$1 ) & Conv2D(3$\times$1 / 1$\times$1 / 2$\times$1 / 2$\times$1 )& Conv2D(3$\times$1 / 1$\times$1 / 3$\times$1 / 3$\times$1 ) \\
		\cline{2-4}
	 &\multicolumn{3}{c|}{Conv2D (1 $\times$ 1 / 1$\times$1)}\\
	 \cline{2-4}
		&Conv2D (1 $\times$ 1 / 1$\times$1)&Conv2D (1 $\times$ 1 / 1$\times$1)&{Conv2D (1 $\times$ 1 / 1$\times$1)} \\
	 \cline{1-4}
		\bf{Recognition and Segmentation} &Conv1D (3 $\times$ 3 / 1$\times$1 / 1$\times$1)& \multicolumn{2}{c|}{Conv1D (3 $\times$ 3 / 1$\times$1 / 1$\times$1)}\\
		\hline
	\end{tabular}
\begin{tablenotes}
	\footnotesize
	\item[1] (3$\times$1 / 1$\times$1 / 1$\times$1 / 1$\times$1 ) represents  kernel size, stride, padding, and dilation in order. 
\end{tablenotes}
\end{threeparttable}
}
\end{table*}
\subsection{Model}
The MTHARS approach consists of the SK network\cite{gao2021deep} (as a backone network), the Windows Generate module and the recognition and segmentation module, as shown in Fig. \ref{segmentmodel}. The parameters are presented in Table \ref{network description}. First, the feature sequence is generated via the backone network. Then, the Windows Generate module generates a certain number of windows according to the length of the feature sequence. The recognition and segmentation module predicts the activity class contained in each window and the offset between the window and the truth activity bounding box. Then, a non-maximum suppression algorithm is used to retain the final satisfactory windows.
\begin{algorithm}[htb]
	\caption{Concatenate algorithm}\label{Prediction}
	\renewcommand{\algorithmicrequire}{\textbf{Input:}}
	\renewcommand{\algorithmicensure}{\textbf{Output:}}
	\begin{algorithmic}[1]
		\REQUIRE   
		Predicted Activity boundary offset $\mathbf{F}$ = \{$\mathbf{f}^{x}_{i}$, $\mathbf{f}^{l}_{i}$\}$_{i=1}^{N}$ \\ 
		Window $\mathbf{W}$ = \{$\mathbf{w}^{x}_{i}$, $\mathbf{w}^{l}_{i}$\}$_{i=1}^{N}$\\
		Predicted activity probability of the activity boundary \textbf{c} = $\{c_{i}$\}$_{i=1}^{N}$,\\
		Predicted activity class of the activity boundary \textbf{y} = $\{y_{i}$\}$_{i=1}^{N}$ , \\
		
		\ENSURE
		A set of activity boundaries ${\mathcal{T}}$ = \{(\textit{s$_i$}, \textit{e$_i$})\}$_{i=1}^K$ associated with activity class \textbf{a} = \{${a_i}$\}$_{i=1}^K$.  \\
		
		\STATE Remove redundant windows via the NMS algorithm \\	
		\STATE Calculate Eq. (\ref{predictionActivityCx})(\ref{predictionActivityL}) to obtain Predicted activity boundaries $\mathbf{P}$ = \{$\textit{P}_{i}$\}$_{i=1}^L$
		\STATE a$_{1}$ $\leftarrow$ The first activity class in the first segment 
		\STATE s$_{1}$ $\leftarrow$ The starting position of the first activity in the first segment 
		\STATE e$_{1}$ $\leftarrow$  The ending position of the first activity in the first segment \\
		\STATE Initial index i
		\FOR{$j = 1$ to \textit{L} }
		\IF{a$_{1}$ $\neq$ a$_{j}$}
		\STATE Add a$_{1}$ to ${a_i}$
		\STATE Add (s$_{1}$, e$_{1}$) to ${\mathcal{T}_{i}}$\\
		\STATE a$_{1}$ $\leftarrow$ a$_{j}$
		\STATE s$_{1}$ $\leftarrow$ e$_{1}$ + 1
		\STATE i $\leftarrow$ i + 1
		\ENDIF
		\STATE e$_{1}$ $\leftarrow$ boundary length of $\textit{P}_{j}$  + s$_{1}$ 
		\ENDFOR
		\IF{$\mathcal{T}$ is Empty}
		\STATE Add a$_{1}$ to ${a_i}$
		\STATE Add (s$_{1}$, e$_{1}$) to ${\mathcal{T}_{i}}$ 
		\ENDIF
	\end{algorithmic}
\end{algorithm}
\subsection{Recognition and Segmentation Module}
	The module is a network comprised of two branches for offset and class prediction. For the class prediction branch, suppose the number of activity classes is \textit{k}, and each generated window has \textit{k}+1 classes, where class 0 represents the background. The length of the feature sequence is \textit{n}. When centered on each unit of the feature sequence generating \textit{m} windows, a set of \textit{n} $\times$ m $\times$ 2 windows need to be classified. If we choose a fully connected layer, it is easy to generate too many model parameters; therefore, we choose a convolutional layer.
	Specifically, the class prediction branch uses a convolutional layer that does not change the length of the feature sequence. By this means, the output and input coordinates correspond to each other in the length of the feature sequence. Consider the input and output for the same \textit{x} coordinate: The class predictions of all the windows centered on the x-coordinate of the input feature sequence are stored in the channel at the x-coordinate of the output feature sequence. For class prediction, there are \textit{m}(\textit{k}+1) output class channels, with the index of the channel \textit{i(k+1)+j (0 $\leq$ j $\leq$ k)} representing the class index \textit{j} predictions for the window index \textit{i}. The design of the offset prediction is similar to that of the class prediction. The only difference is that we predict two offsets for each window instead of \textit{k}+1 activity classes.

\subsection{Training Model}
During the forward propagation of the model, we generate multiscale windows and predict their activity classes and window offsets. The truth activity class and truth offset are labeled on each generated window using the label information. Finally, we calculate the loss based on the predicted classes and offsets with the true classes and offset values, respectively. The offset loss (\ref{l1 loss}) is calculated by comparing the ground truth offset \textit{f} = (\textit{f$^x$}, \textit{f$^l$}) with the predicted offset \textit{\textit{$\hat{f}$}} = (\textit{f$^x$}, \textit{f$^l$}) using Smooth$_{L_{1}}$ loss.
\begin{equation}
	L_{loc}(\textit{f}, \textit{$\hat{f}$}) = \sum_{i\in\{\textit{x}, \textit{l}\}} Smooth_{L1}(f^{i} - \hat{f}^{i})
	\label{l1 loss}
\end{equation}
in which 

\begin{equation}
	Smooth_{L1}(x) = \left\{
	\begin{aligned}
		0.5(x){^2},  if |x|<1. \\
		|x|-0.5,  otherwise.
	\end{aligned}
	\right.\label{smoothl1}
\end{equation}
The class loss is calculated by the window labeled activity class $\mathbf{a}$ with the predicted class $\mathbf{\hat{a}}$ using cross-entropy loss (\ref{category loss}), in which \textit{n} represents the sample number.
\begin{equation}
	L_{conf}(\mathbf{a}, \mathbf{\hat{a}}) = -\sum_{i=1}^{n}a_i log(\hat{a_i})\label{category loss}
\end{equation}

Due to the limited number of activities, there will be many windows that are not matched with any of the activities after matching the window with the truth activity boundary, and these windows are referred to as negative samples. We sort all the negative samples in descending order according to the activity class probability and select a certain number of negative samples with a higher class probabilities to participate in the class loss calculation, with the number of negative samples to the number of positive samples in the ratio of 3:1. If all the negative samples are involved in the training, it will make the network training process tend towards the negative samples.

Finally, we multiply the weight $\alpha$ by the class loss (conf) plus the weight $\beta$ by the localization loss (loc).
\begin{equation}
	L(a,\hat{a}, f, \hat{f}) = \frac{1}{N}(\alpha L_{conf}(a, \hat{a})+\beta L_{loc}(f, \hat{f}))\label{loss}     .
\end{equation}

\subsection{Prediction}
   Since the length of the activity input to the network is fixed, we input a fixed length activity data stream at a time to recognize the start and end positions of each activity in each segment. Finally, we concatenate all the segments according to the activity class to obtain the start and end positions of each activity in the whole activity data stream. The concatenation algorithm is described in algorithm \ref{Prediction}.
\begin{table*}[htbp]
	\renewcommand\arraystretch{1.5}
	\centering 
	\caption{\label{datasetsTable} SIMPLE DESCRIPTION OF PUBLIC HAR DATASETS. }
	\begin{threeparttable}
		\begin{tabular}{lcccccccc}
			\toprule
			\diagbox{\bf{Attribute}}{\bf{Dataset}}& SKODA & HCI &PS & WISDM &UCI & 
			OPPORTUNITY&PAMAP2&UNIMIB SHAR \\
			\midrule
			Type & AG & AG & ADL &ADL& ADL &ADL&ADL&ADL \\
			Subject & 1 & 1 & 4 &29&30& 4 &9& 4\\
			Rate &  96HZ &96HZ & 50HZ &20HZ &50HZ&30HZ&33.3HZ&30HZ \\
			WindowSize &1s &1s & 2s&10s&2.56s&1s&1s&1s\\
			Activity Categories &10&8&6&6&6&18&18&17\\
			Sample & 696975 & 7352 & 161959 &1,098,208&748406 & 701366&2872533&11771 \\
			\bottomrule
		\end{tabular}
		\begin{tablenotes}
			\footnotesize
			\item[1] AG denotes gesture activity and  ADL denotes activity of daily life.
		\end{tablenotes}
	\end{threeparttable}
\end{table*}
\begin{figure*}[htbp]
	\centering
	\subfloat[]{\includegraphics[width=1.9in]{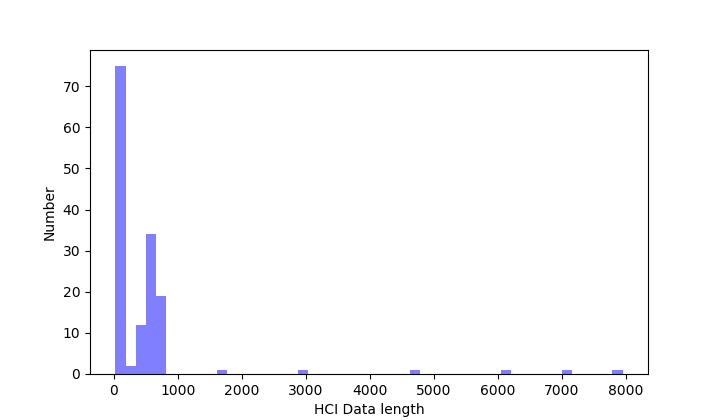}%
		\label{fig_first_case}}
	\subfloat[]{\includegraphics[width=1.9in]{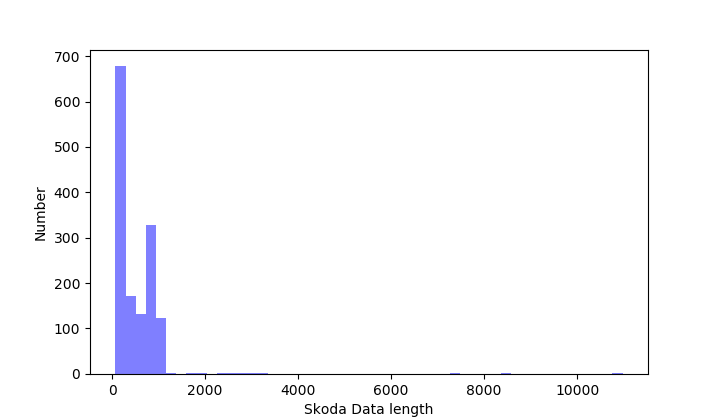}%
		\label{fig_second_case}}
	\subfloat[]{\includegraphics[width=1.9in]{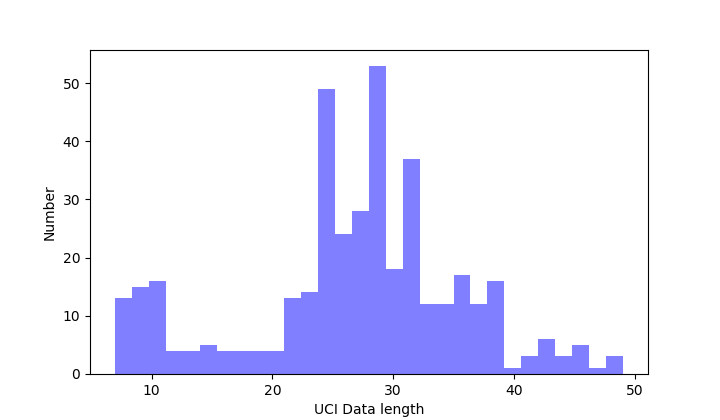}%
		\label{fig_third_case}}
	\subfloat[]{\includegraphics[width=1.9in]{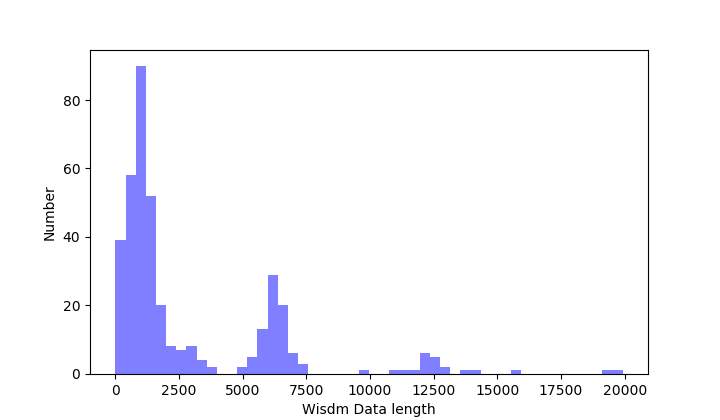}%
		\label{fig_third_case}}
	\hfil
	\subfloat[]{\includegraphics[width=1.9in]{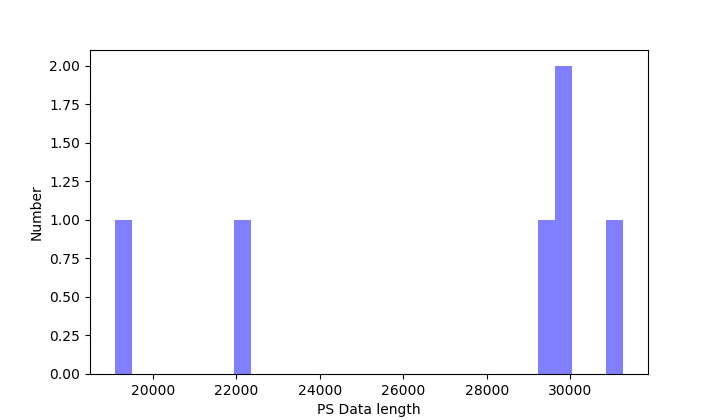}%
		\label{fig_third_case}}
	\subfloat[]{\includegraphics[width=1.9in]{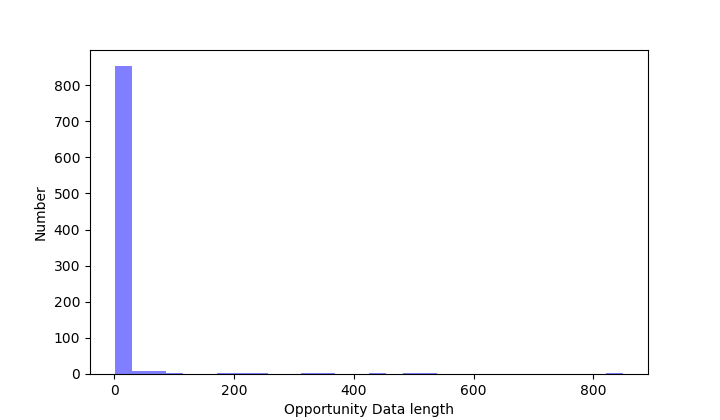}%
		\label{fig_third_case}}
	\subfloat[]{\includegraphics[width=1.9in]{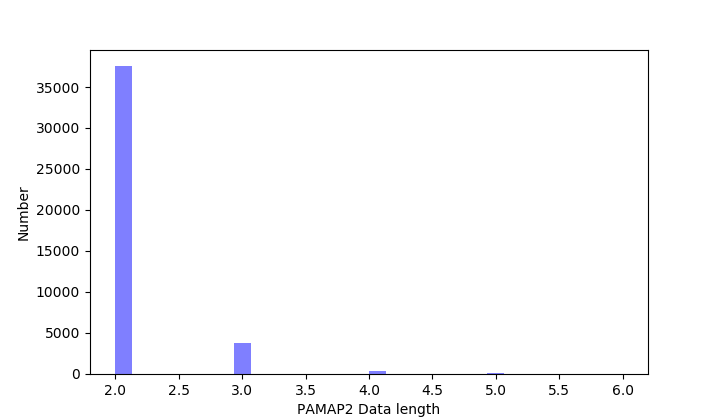}%
		\label{fig_third_case}}
	\subfloat[]{\includegraphics[width=1.9in]{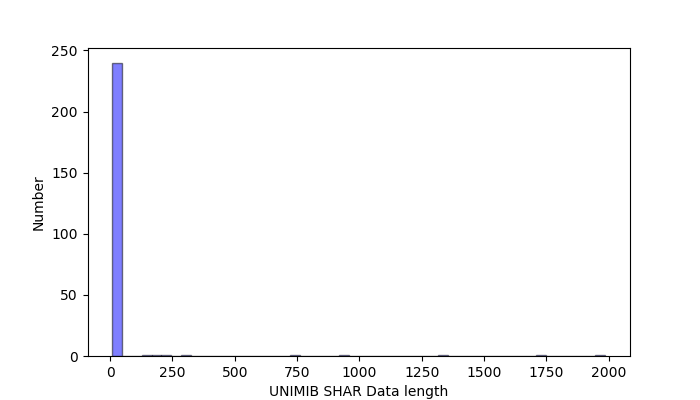}%
		\label{fig_third_case}}
	\caption{Activity length distribution of benchmarking datasets}
	\label{datasets distribution figure}
\end{figure*}

\section{EXPERIMENTS AND RESULTS}
In this section, we detail our experiments and analyze the experimental results. We conduct an extensive experiment on eight benchmarking datasets to evaluate the effectiveness and scalability of our proposed MTHARS.
\subsection{Datasets}

We use eight publicly benchmark HAR datasets to assess the effectiveness of our system in daily activity segmentation and recognition. To make a fair comparison with other researchers, we select the same parameter values adopted by them. The specifics of the eight benchmark data we use are described in Table \ref{datasetsTable}, and the length interval distribution of benchmarking datasets is depicted in Fig. \ref{datasets distribution figure}.
\begin{itemize}
	\item SKODA Dataset\cite{stiefmeier2007fusion}: A participant wearing 10 USB acceleration sensors in the upper and lower left and right hands, respectively, performed 10 different gestural activities in an automobile repair scenario for 3 hours. Each of the ten gesture actions, such as checking the gasoline tank, was completed more than 70 times.
	\item HCI Dataset\cite{forster2009unsupervised}: Eight accelerometer USB sensors were worn on the lower and upper half of one participant's right arm to perform various gestural activities, such as sketching triangles, squares, and circles. More than 10 freehand movements and over 60 guided gestures were performed.
	\item PS Dataset \cite{shoaib2014fusion}: Four participants recorded their walking, standing, standing, going upstairs, and going downstairs activities. They employed the phone's built-in accelerometer, magnetometer, and gyroscope by placing a mobile phone in four different locations on the body: pants pocket, waistband, right arm, and right wrist.
	\item  WISDM Dataset\cite{kwapisz2011activity}: The data were obtained by 29 participants using phones with triaxial acceleration sensors placed in their pant pockets with a sampling frequency of 20 Hz. Walking, strolling, walking up stairs, walking down stairs, standing motionless and standing up were among the six daily activities undertaken by each participant. The mean value of the column was used to fill in the missing values in the dataset.
	\item UCI Dataset\cite{anguita2013public}: The dataset consisted of 30 participants aged 19 to 48 years performing a series of daily activities. Each participant wore a Samsung Galaxy S2 smartphone around their waist and collected sensor data in 9 dimensions using the phone's built-in acceleration, gyroscope, linear acceleration and 3-axis angular velocity sensors and performed 6 different activities of daily living (walking, walking up stairs, walking down stairs, standing still, standing, and lying down).
	\item OPPORTUNITY Dataset\cite{anguita2013public}: IMU sensors were placed on 12 different places of each volunteer's body, and they were asked to repeat a sequence of 17 morning activities in the kitchen, such as opening the refrigerator door, closing the refrigerator door, opening the drawer, closing the drawer, and so on. Interpolation was performed to fill in missing values in the dataset.
	\item PAMAP2 Dataset\cite{reiss2012introducing}: Each of the nine participants wore an IMU on their chest, hands, and ankles, which collected acceleration, angular velocity, and magnetic force sensor information. Each participant was required to complete 12 mandatory activities, such as lying, standing, and walking up and down stairs, as well as six elective activities, such as watching television, driving, and playing ball. Due to interference generated during the activity transformation, the start and the last 10 s of each activity were eliminated to reduce the noise in the data for each type of activity.
	\item UNIMIB SHAR Dataset\cite{micucci2017unimib}: The dataset was collected by researchers from the University of Milano-Bicocca. The researchers equipped each of the 30 volunteers with Bosh
	BMA220 sensor Samsung cell phones and placed them in the participants' left or right pockets to collect activities of daily life (running, going upstairs, standing, walking, etc.) as well as different falling activities. Each activity was repeated 3 to 6 times.
\end{itemize}
\subsection{Evaluation metric}
To assess the accuracy of the activity segmentation, we use the Normalized Edit Distance
(NED)\cite{aminikhanghahi2019enhancing}. NED uses the Levenshtein distance to measure how far apart the predicted activity sequence (\textit{$\hat{T}$}) is from the truth activity sequence (\textit{T}), calculated by the smallest operation that makes the two sequences equivalent, and NED is defined as follows.

	\begin{equation}
	NED = \frac{lev(\hat{T}, T)}{length\, of\, T} 
\end{equation}

\begin{equation}
	lev(i,j) = \left\{
	\begin{aligned}
		\max(i,j)  \min(i,j) = 0 \\
		min = \left\{
		\begin{aligned}
			lev(i-1,j)+1 \\
			lev(i,j-1)+1\\
			lev(i-1,j-1)+1_{i \neq j}.
		\end{aligned}
		\right.
	\end{aligned}
	\right.\label{lev equation}
\end{equation}
Equation (\ref{lev equation}) is the smallest number of operation steps required to make the predicted sequence and the truth sequence equal, that takes into account three different ways to make the two sequences equal, namely, removing an element from the sequence, inserting a new one, and changing the sequence's label directly. This metric allows us to calculate the difference between the predicted activity boundary and the true activity boundary.

For an evaluation of activity recognition accuracy, since the activity classes in human activity data are mostly unbalanced, using classification accuracy is not an appropriate criterion to evaluate the classification performance\cite{ordonez2016deep}. Therefore, we apply \textit{F$_{1}$} to evaluate the effectiveness of activity classification.

\begin{equation}
	F_{1} = 2\sum\frac{N_{c}}{N_{total}}\frac{P\times R}{P + R} 
\end{equation}

N$_{c}$ represents the number of class c in all the samples, and N$_{total}$ denotes the number of all the samples. \textit{P} and \textit{R} are calculated from the set of all positive classes, defined as below.

\begin{equation}
	P = \frac{\sum_{i=1}^n TP_i}{\sum_{i=1}^n TP_i + \sum_{i=1}^n FP_i} 
\end{equation}

\begin{equation}
	R = \frac{\sum_{i=1}^n TP_i}{\sum_{i=1}^n TP_i + \sum_{i=1}^n FN_i} 
\end{equation}
The activity class in the dataset is represented by \textit{i}. \textit{TP}$_{i}$ represents the true positive class, \textit{FP$_{i}$} denotes the false-positive class, and \textit{FN}$_{i}$ indicates the false-positive class.

\subsection{Static Sliding-Window Segmentation}
In this section, to investigate the effect of the static window segmentation methods, we use time-based sliding windows for experiments. As mentioned above, a time-based sliding window referrs to segmenting the activity data stream into the same length of time. The segmented data are randomly split into training and testing with a ratio of 70\%, 30\%. We used the commonly used time duration \textit{t} from 1 to 10 s.
\begin{table}[htbp]
	\renewcommand\arraystretch{1.5}
	\centering 
	\caption{\label{sliding window f1 result}F$_{1}$ Score of STATIC TIME-BASED SLIDING WINDOW.}
	\begin{tabular}{lcccccc}
		\toprule
		\bf{Datasets}&\bf{Methods}& \textit{t} = 1s &  \textit{t} = 2s  &  \textit{t} = 3s  &\textit{t} = 5s & \textit{t} = 10s\\
		\midrule
		\multirow{5}{*}{\bf{SKODA}}&NB&0.6351&0.6691&0.6027 & \bfseries{0.8292}&0.8123\\
		&DT&0.8540&0.8928&0.9202&0.9364&\bf{0.9657}\\
		&SVM&0.8496&0.8830&\bf{0.9461}&0.9065&0.9340\\
		&LSTM&0.8656&0.8756&0.9122&\bf{0.9336}&\bf{0.9336}\\
		&GRU&\bf{0.8605}&0.8367&0.7932&0.7563&0.8435\\
		\cline{1-7}
		\multirow{5}{*}{\bf{HCI}}&NB&0.6039&0.6840&0.648 1& 0.5784&\bf{0.8099}\\
		&DT&0.6581&0.7552&0.8441&\bf{0.8693}&0.8249\\
		&SVM&0.5323&0.5976&0.6771&\bf{0.8700}&0.7631\\
		&LSTM&0.7523&0.7124&\bf{0.7726}&0.6950&0.6269\\
		&GRU&0.7474&0.7193&\bf{0.7493}&0.7109&0.7944\\
			\cline{1-7}
		\multirow{5}{*}{\bf{PS}}&NB&0.6378&\bf{0.6648}&0.6064& 0.5961&0.5923\\
		&DT&0.9340&0.9332&\bf{0.9481}&0.9436&\bf{0.9481}\\
		&SVM&0.5185&\bf{0.5223}&0.5180&0.5180&\bf{0.5228}\\
		&LSTM&0.7277&\bf{0.7777}&\bf{0.7777}&\bf{0.7777}&0.7759\\
		&GRU&\bf{0.8318}&0.8317&\bf{0.8318}&0.8283&0.8283\\
			\cline{1-7}
		\multirow{5}{*}{\bf{WISDM}}&NB&\bf{0.7516}&0.7155&0.6763& 0.6679&0.6380\\
		&DT&0.8097&0.8829&0.9161&\bf{0.9604}&0.8821\\
		&SVM&\bf{0.8275}&0.8244&0.8191&0.811&0.8106\\
		&LSTM&0.8333&0.8946&0.8900&0.9011&\bf{0.9090}\\
		&GRU&0.9365&0.953&\bf{0.9562}&0.95214&0.9217\\
			\cline{1-7}
		\multirow{5}{*}{\bf{UCI}}&NB&\bf{0.6670}&0.6704&0.5283&0.2061&0.3059\\
		&DT&\bf{0.5984}&0.5428&0.5723&0.5130&0.5595\\
		&SVM&0.7374&0.7195&0.7623&0.6842&\bf{0.7778}\\
		&LSTM&0.9366&0.9158&0.8887&0.9210&\bf{0.9490}\\
		&GRU&\bf{0.8114}&0.7194&0.5283&0.6935&0.6555\\
		\bottomrule
	\end{tabular}
\end{table}

We select several different classification models to investigate the impact of time-based static window segmentation. The results of the time-based static segmentation technique for each \textit{t} are shown in Table \ref{sliding window f1 result}.

As Table \ref{sliding window f1 result} demonstrates, on the SKODA dataset, DT and LSTM achieved the best F$_{1}$ values of 0.9657 and 0.9336, respectively, at \textit{t}=10 s, while NB and SVM achieved the best F$_{1}$ values of 0.8292 and 0.9336, respectively, at \textit{t}=5 s. The optimal window lengths chosen are different for the different methods. On the WISDM dataset, we found that NB and SVM achieved 0.7516 and 0.8275 F$_{1}$ values at \textit{t}=1 s, while for DT and LSTM, the best results are achieved at \textit{t}=5 and \textit{t}=10 s. The longer the length of the window for the same dataset, the more information it contains, but it may fail to recognize the active transition points. Instead, the smaller the length of the window may capture the activity transition points but may misclassify them because the window does not contain enough key information.

From the above experimental results, we do not compare which method performs better at classification because we tested specific parameters on the five datasets. We investigate whether the length of the window affects the F$_{1}$ score of the method. We conclude that the optimal window length is different for different methods on the same dataset and the optimal window length chosen for the same method on different datasets also varies due to the variable duration of the activity on different datasets.
\begin{figure}[htpb]
	\centering
	\includegraphics[width=0.5\textwidth]{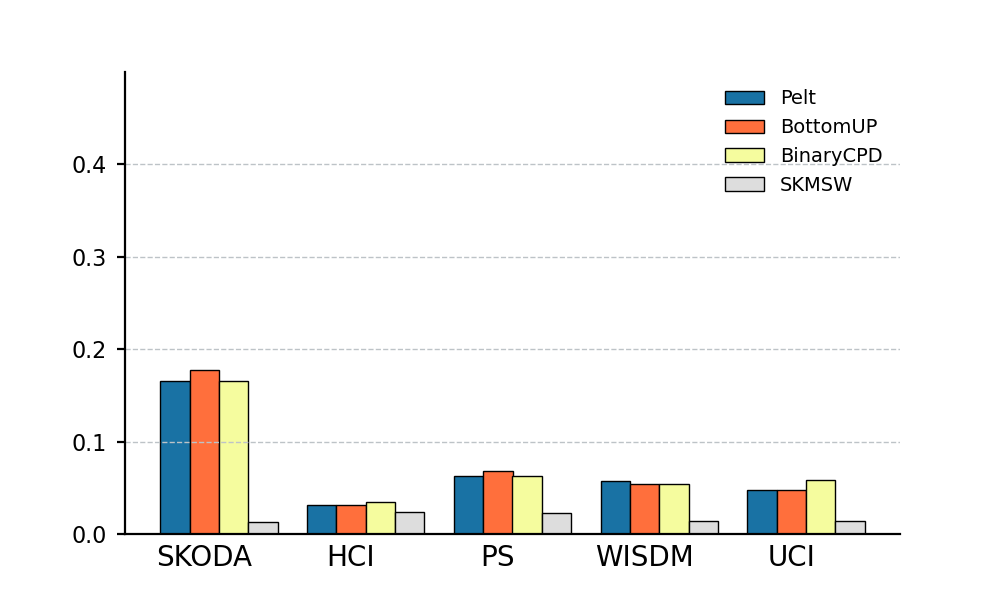}
	\caption{NED performance on the five benchmarking datasets}
	\label{NED result figure}
\end{figure}
\subsection{Dynamic Segmentation}
As shown in the above experiments based on static window segmentation, a suitable window length can improve the performance of classification, but in real life, choosing an optimal window length is difficult. Therefore, in this section, we use dynamic segmentation based on sensor data. We do not want to compare the segmentation performance with the following benchmark because it is slightly unfair. We attempt to investigate the relationship between segmentation accuracy and classification accuracy on the same dataset. The benchmark method uses dynamic segmentation followed by feature extraction. The feature extraction is fixed, and our method performs feature extraction followed by segmentation. The F$_{1}$ score of the segmentation methods is shown in Table \ref{dynamic segmentation result}.
\begin{table}[htbp]
	\renewcommand\arraystretch{1.5}
	\centering 
	\caption{\label{dynamic segmentation result} F$_{1}$ Score of DYNAMIC SEGMENTATION.}
	\begin{tabular}{lccccc}
		\toprule
		\diagbox{\bf{Methods}}{\bf{Datasets}}& SKODA & HCI & PS &WISDM & UCI\\
		\midrule
		Dynp\cite{guedon2013exploring}&0.8858&0.8751&0.9663 &0.8711&0.9352\\
		BottomUP\cite{keogh2001online} &0.8661&0.8750&0.9536&0.8807 &
		0.9353\\
		BinaryCPD\cite{fryzlewicz2014wild} &0.8826&0.8450&0.9601&0.8859&
		0.9151\\
		MTHARS &\bf{0.9648}&\bf{0.9479}&\bf{0.9733}&{\bf{0.9872}} &
		{\bf{0.9632}}\\
		\bottomrule
	\end{tabular}
\end{table}

As shown in Table \ref{dynamic segmentation result}, the F$_{1}$ values of all the methods exceed 90\% on both the PS and UCI datasets. However, the F$_{1}$ values of all the methods on the HCI dataset are lower than the accuracy of the remaining four datasets. 
Our proposed method is superior to the dynamic segmentation-based method, possibly because our method better identifies the starting and ending positions of the activity, which can improve the classification performance.
 
To verify our hypothesis, we investigated the relationship between the performance of the classification and the segmentation accuracy. The overall results of the segmentation performance are shown in Fig. \ref{NED result figure}. A smaller NED value means that the position of the segmented activity sequence is closer to the position of the real activity sequence.
\begin{table*}[htbp]
	\renewcommand\arraystretch{1.5}
	\centering 
	\caption{\label{activity recognition result}F$_{1}$ Performance on eight data sets.}
	\begin{tabular}{lcccccccc}
		\toprule
		\diagbox{\bf{Methods}}{\bf{Dataset}} & SKODA & HCI & PS & WISDM & UCI &OPPORTUNITY&PAMAP2&UNIMIB SHAR\\
		\midrule
		SK\cite{gao2021deep} & 0.9510 & 0.9377 & 0.9574 & 0.9725 & 0.9558&0.9074&0.9338&0.7463  \\
		MTHARS & {\bf{0.9632}} & 0.9524 & {\bf{0.9721}} & {\bf{0.9877}} &
		{\bf{0.9723}} &{\bf{0.9213}}&{\bf{0.9480}}&{\bf{0.7571$  $}}\\
		\hline
		Other Reachers&0.958*\cite{ordonez2016deep}&	
		&  
		& 0.9263\cite{xia2020lstm} 
		&0.9302\cite{cruciani2020feature}&0.915\cite{ordonez2016deep}&0.9248\cite{tang2022triple}&0.7538\cite{tang2022triple}
		\\
		& 0.928\cite{abedin2021attend}&  &
		&0.949\cite{varamin2018deep} &
		0.9585\cite{xia2020lstm}&0.9263\cite{xia2020lstm}
		&0.9116\cite{wan2020deep}&\\
		&0.924\cite{guan2017ensembles} &	 &
		&  0.9720\cite{Khan2021}  &
		0.9545\cite{tong2022novel}&0.849\cite{varamin2018deep}&0.908\cite{abedin2021attend}\\
		&0.916\cite{yao2018efficient} &
		& 
		&& 0.9660\cite{tang2022triple} &0.746\cite{abedin2021attend}&0.854\cite{guan2017ensembles} \\	
		&
		&&&&0.9537\cite{Khan2021}&0.726\cite{guan2017ensembles} &0.9303\cite{teng2020layer}\\
		&     &&&&0.9293\cite{wan2020deep}&0.8058\cite{teng2020layer}\\
		
		\bottomrule
	\end{tabular}
\end{table*}
From Fig. \ref{NED result figure}, we can see that our method achieves the lowest NED values, and the classification performance is more accurate. on the HCI dataset, we find that the NED values of Dynp and BottomUP are similar, and their classification results are also close to each other. On the PS dataset, the NED values of Dynp and BinaryCPD are lower than those of the BottomUP method, and the F$_{1}$ score of their classifications is higher than that of the BottomUP classification. On the WISDM dataset, the BottomUP and BinaryCPD methods have higher classification results than Dynp, and their NED values are lower than those of the Dynp method. 

From the above experimental results, it is indicated that precision in obtaining the start and end positions of the activities can improve the performance of the recognition activities. If data segmentation is considered as a preprocessing process, errors in data segmentation may be propagated to the later steps.

\subsection{Activity Recognition}
In this scenario, we investigate the classification performance of our method compared with SK\cite{gao2021deep}. For the baseline methods, we leverage the segmentation methods mentioned in their papers (sliding window methods are applied). The datasets were randomly divided into training and testing sets according to the experimental benchmark, where the training and testing sets in the PAMAP2 dataset were in a ratio of 80\% and 20\%, respectively, and the proportions of the training and testing sets in the remaining seven datasets were 70\% and 30\%, respectively. The missing values in the dataset were linearly interpolated or repeated from the previous values, and data preprocessing was performed by normalization. 

\begin{table*}[htbp]
	\renewcommand\arraystretch{1.5}
	\centering 
	\caption{\label{activity recognition accuracy result}Accuracy Performance on eight datasets.}
	\begin{tabular}{lcccccccc}
		\toprule
		\diagbox{\bf{Methods}}{\bf{Dataset}}& SKODA & HCI & PS &WISDM & UCI &OPPORTUNITY&PAMAP2&UNIMIB SHAR\\
		\midrule
		SK\cite{gao2021deep}&0.9586&0.9341&0.9649 &0.9751 & 0.9406&0.9014&0.9380&0.7589  \\
		MTHARS &\bf{0.9648}&\bf{0.9479}&\bf{0.9733}&{\bf{0.9872}} &
		{\bf{0.9632}} &{\bf{0.9153}}&{\bf{0.9450}}&{\bf{0.7648$  $}}\\
		\bottomrule
	\end{tabular}
\end{table*}

Table \ref{activity recognition result} reports the experimental results of MTHARS and the baselines. To further evaluate the generality of our method, we ran the SK code using the same evaluation criteria (as shown in Table \ref{activity recognition accuracy result}). The result shows that our proposed method outperforms the state-of-the-art performance.

1) Improvements on the SKODA dataset: Compared with SK\cite{gao2021deep}, our method obtains a 0.22 \% higher F$_{1}$ score. We also compare the MTHARS method with previously published results. To the best of our knowledge, the best result is \cite{ordonez2016deep}, and the MTHARS method obtains a 0.52\% higher F$_{1}$ score.

2) Improvements on the HCI dataset: The results show that our method outperforms SK\cite{gao2021deep} by obtaining a 1.47 \% higher F$_{1}$ score.

3) Improvements on the PS dataset: When compared with SK\cite{gao2021deep}, our method obtains a 1.47 \% higher F$_{1}$ score.

4) Improvements on the WISDM dataset: Table \ref{activity recognition result} shows that the MTHARS outperforms the SK\cite{gao2021deep} method by a 1.52\% higher F$_{1}$ score. Moreover, we can observe that the MTHARS attains a 1.57\% higher F$_{1}$ score than the rest of the published results \cite{Khan2021}.

5) Boost on the UCI dataset: we compare the MTHARS with the SK\cite{gao2021deep} method on the test set. As shown in Table \ref{activity recognition result}, the MTHARS obtains a 1.65\% improvement. We compare the MTHARS to the recently published results on the UCI dataset. According to our investigation, the best result thus far is 0.9660\cite{tang2022triple}, and the MTHARS obtains a 0.63\% higher F$_{1}$ score.

6) Improvements on the OPPORTUNITY dataset: We still use the same method for comparison. In Table \ref{activity recognition result}, we show our experimental results that outperform the SK method with a 1.39\% improvement on the test dataset. Then, we compare the results with the currently published studies as follows: \cite{ ordonez2016deep},\cite{varamin2018deep},\cite{guan2017ensembles},\cite{teng2020layer}. We can see from Table \ref{activity recognition result} that the MTHARS obtains 0.63\%, 7.23\%, 19.53\% and 11.55\% improvements compared to them, respectively. Compared to \cite{xia2020lstm}, the MTHARS results are closer to it.
\begin{figure}[htpb]
	\centering
	\subfloat[]{\includegraphics[width=0.45\textwidth]{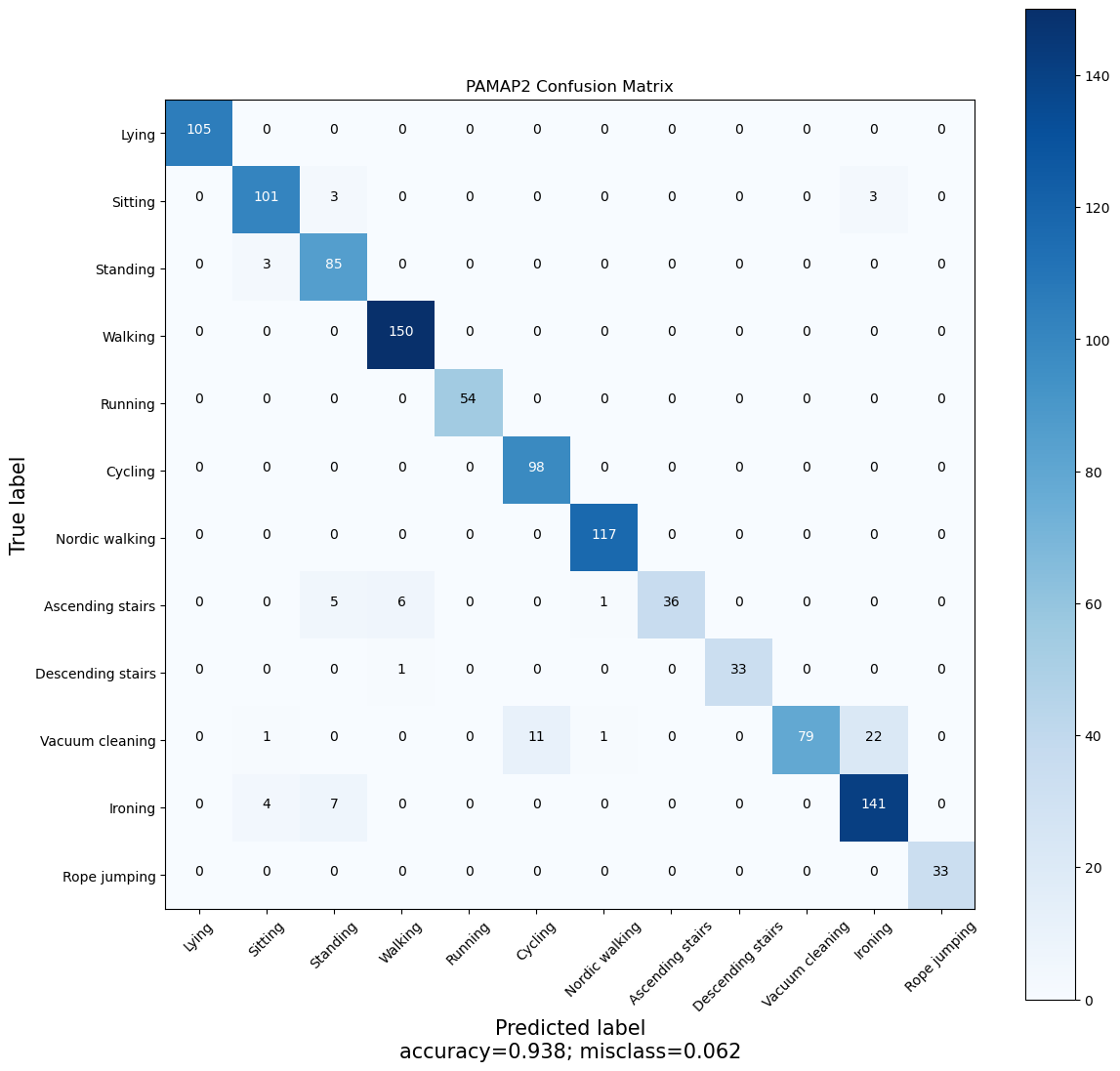}%
		\label{pamap_baseline}}
	
	\subfloat[]{\includegraphics[width=0.45\textwidth]{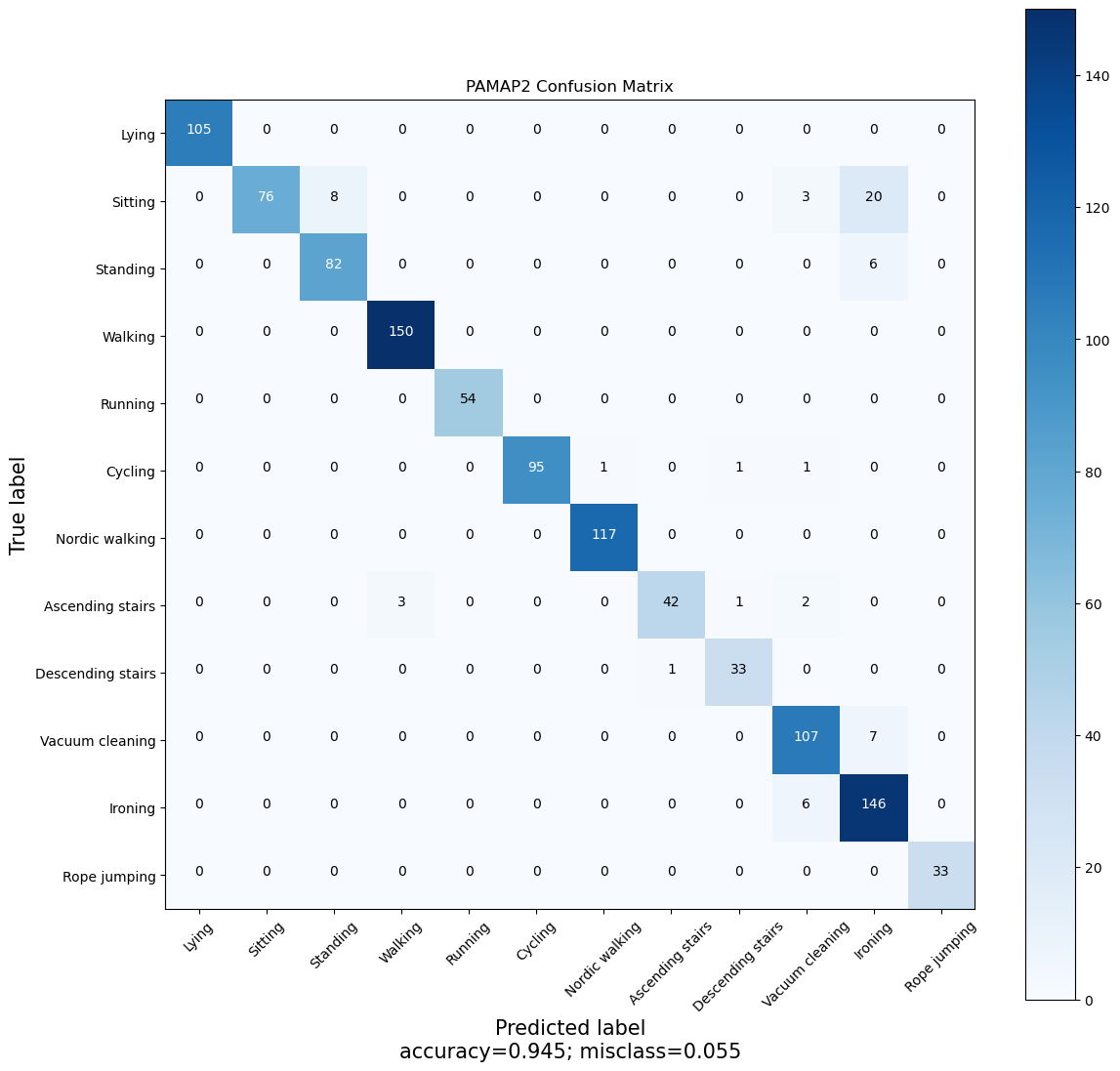}%
		\label{pamap_ssms}}
	\caption{The confusion matrices on the PAMAP2 dataset between SK\cite{gao2021deep} and MTHARS. (a) SK, (b) MTHARS}
	\label{confusion_pamap2}
\end{figure}

7) Boost on the PAMAP2 dataset: According to Table \ref{activity recognition result}, the F$_{1}$ score of MTHARS surpasses that of SK on this dataset. Compared with the recently published results \cite{guan2017ensembles}, \cite{tang2022triple }, \cite{teng2020layer}, the MTHARS obtains a 9.32\%, 2.32\% and 1.77\% improvement. The confusion matrices have
been computed in Fig.\ref{confusion_pamap2} to visually show the performance
improvements.

8) Boost on the UNIMIB SHAR dataset: We investigated the impact of our method on the experimental results. Table \ref{activity recognition result} shows that MTHARS receives a 1.08\% boost compared to SK. We also compare MTHARS with other published research. From our investigation, the best value on the UNIMIB SHAR dataset is \cite{tang2022triple}, with which the MTHARS obtains a 0.33\% improvement.

From the above results, we concluded that the MTHARS method, performing activity segmentation and recognition simultaneously, is able to improve the accuracy of activity classification. Segmentation of the activity by employing multiscale windows can provide an accurate boundary for recognizing the activity. Moreover, the two tasks can mutually facilitate each other.
\subsection{Ablation experiments}
In this subsection, we conduct ablation experiments on the benchmarking datasets to investigate the effectiveness of the approach. During our experiments, we found that the class loss and offset loss weights and the scale \textit{s} of the window are two essential settings.

First, the weights \textit{$\alpha$} and \textit{$\beta$} represent the different proportions of class loss and offset loss in the loss calculation. For simplicity, we first set the scale \textit{s} of the window to 2 and 3.
To investigate how they affect the experimental classification results, we considered different values of the weights \textit{$\alpha$} and \textit{$\beta$} on the OPPORTUNITY and WISDM datasets. We considered that when \textit{$\alpha$}=1
\textit{$\beta$}=1, \textit{$\alpha$}=1 \textit{$\beta$}=2, \textit{$\alpha$}=1 \textit{$\beta$}=3, \textit{$\alpha$}=2
\textit{$\beta$}=1. The detailed results of the experiment are shown in Table \ref{different weight setting}. From the table, we can see that the best result on the WISDM dataset is the combination of \textit{$\alpha$}=2 \textit{$\beta$}=3. In addition, when \textit{$\alpha$}=2, \textit{$\beta$}=3 is very similar to the result when \textit{$\alpha$}=1 \textit{$\beta$}=1. The value of 0.9796 when \textit{$\alpha$}=1 \textit{$\beta$}=2 is very close to the result of 0.9783 when \textit{$\alpha$}=2 \textit{$\beta$}=1. On the OPPORTUNITY dataset, we can see from the table that \textit{$\alpha$}=1 \textit{$\beta$}=3 is close to the results for two different combinations of \textit{$\alpha$}=2 \textit{$\beta$}=3, \textit{$\alpha$}=1 \textit{$\beta$}=1.
\begin{table}[htbp]
	\centering 
	\caption{\label{different weight setting}Activity classification F$_{1}$ value with different weight settings on different data sets}
	\begin{tabular}{lcc}
		\toprule
		Model &OPPORTUNITY& WISDM \\
		\midrule
		SK\cite{gao2021deep}&0.9074 &0.9725\\
		\midrule
		$\alpha$=1,$\beta$=1 &{\bf{0.9213}}& 0.9877\\
		$\alpha$=1,$\beta$=2 &0.9060& 0.9796 \\
		$\alpha$=1,$\beta$=3 &0.9174& 0.9874 \\
		$\alpha$=2,$\beta$=1 &0.9075& 0.9783 \\
		$\alpha$=2,$\beta$=3 &0.9154& {\bf{0.9881}} \\
		\bottomrule
	\end{tabular}
\end{table}

Next, we seek to better understand the effect of different sizes and numbers of scale \textit{s} on the experimental classification results. Therefore, we set various numbers and scales \textit{s} on the OPPORTUNITY and UCI datasets. We believe that different sizes and numbers of scale \textit{s} generating the number and length of windows are different. Different numbers and lengths of windows produce different offset values from the true activity boundary, which will have an impact on the experimental results. Due to the dataset length limitation, we only set it when there is only one \textit{s}=2, \textit{s}=2, 3, \textit{s}=0.5, 0.3 and \textit{s}=2, 3, 4, respectively. Specifically, the weights $\alpha$ and $\beta$ are set to 1. The detailed results of the experimental classification are shown in Table \ref{different s setting}. We found that when only one \textit{s} is set, the effect is not as good as the classification effect when several different \textit{s} are set. On the OPPORTUNITY dataset, setting the scale too small, such as \textit{s} = 0.5 and 0.3, is close to the classification result of \textit{s}=2. Setting the scale too large, such as \textit{s}=4, is close to the experimental classification result of \textit{s}=0.5, 0.3. The best classification effect is to set the scale to \textit{s}=2, 3. On the UCI dataset, the best result is 0.9723 when \textit{s}=2, 3, 4. The second result is 0.9615 with \textit{s}=2, 3. The results show that combining classification and segmentation using multiple windows of different scales results in better recognition of activities of different lengths.
		\begin{table}[htbp]
		\centering 
		\caption{\label{different s setting}Activity classification F$_{1}$ value for different \textit{s} on different dataset}
		\begin{tabular}{lcc}
			\toprule
			Model & OPPORTUNITY&UCI \\
			\midrule
			SK\cite{gao2021deep}&0.9074 &0.9558\\
			\midrule
			\textit{s}=2 & 0.9138&0.9505 \\
			\textit{s}=0.5,0.3 & 0.9160&0.8928 \\
			\textit{s}=2,3 & \bf{0.9213}&0.9615 \\
			\textit{s}=2,3,4 & 0.9167&\bf{0.9723 }\\
			\bottomrule
		\end{tabular}
	\end{table}

\section{Conclusion and future work}
Accurate segmentation of activities, as well as recognition of activities, is a challenging task. In this paper, we propose a method called MTHARS, which generates multiscale windows to segment activities of different lengths. MTHARS combines activity segmentation and recognition, which not only improves the activity segmentation performance but also improves the accuracy of classification. We evaluated the method on eight benchmark datasets and our method outperformed the state-of-the-art methods. To verify the effectiveness of the method, we conducted several ablation experiments on the benchmark dataset. Future research could be conducted to improve the model's performance on activity boundaries with very short durations.

\bibliographystyle{IEEEtran}
\bibliography{IEEEabrv,references}

\begin{thebibliography}{10}
\providecommand{\url}[1]{#1}
\csname url@samestyle\endcsname
\providecommand{\newblock}{\relax}
\providecommand{\bibinfo}[2]{#2}
\providecommand{\BIBentrySTDinterwordspacing}{\spaceskip=0pt\relax}
\providecommand{\BIBentryALTinterwordstretchfactor}{4}
\providecommand{\BIBentryALTinterwordspacing}{\spaceskip=\fontdimen2\font plus
\BIBentryALTinterwordstretchfactor\fontdimen3\font minus
  \fontdimen4\font\relax}
\providecommand{\BIBforeignlanguage}[2]{{%
\expandafter\ifx\csname l@#1\endcsname\relax
\typeout{** WARNING: IEEEtran.bst: No hyphenation pattern has been}%
\typeout{** loaded for the language `#1'. Using the pattern for}%
\typeout{** the default language instead.}%
\else
\language=\csname l@#1\endcsname
\fi
#2}}
\providecommand{\BIBdecl}{\relax}
\BIBdecl

\bibitem{chen2012sensor}
L.~Chen, J.~Hoey, C.~D. Nugent, D.~J. Cook, and Z.~Yu, ``Sensor-based activity
  recognition,'' \emph{IEEE Transactions on Systems, Man, and Cybernetics, Part
  C (Applications and Reviews)}, vol.~42, no.~6, pp. 790--808, 2012.

\bibitem{chen2011knowledge}
L.~Chen, C.~D. Nugent, and H.~Wang, ``A knowledge-driven approach to activity
  recognition in smart homes,'' \emph{IEEE Transactions on Knowledge and Data
  Engineering}, vol.~24, no.~6, pp. 961--974, 2011.

\bibitem{cook2009ambient}
D.~J. Cook, J.~C. Augusto, and V.~R. Jakkula, ``Ambient intelligence:
  Technologies, applications, and opportunities,'' \emph{Pervasive and Mobile
  Computing}, vol.~5, no.~4, pp. 277--298, 2009.

\bibitem{lara2012survey}
O.~D. Lara and M.~A. Labrador, ``A survey on human activity recognition using
  wearable sensors,'' \emph{IEEE communications surveys \& tutorials}, vol.~15,
  no.~3, pp. 1192--1209, 2012.

\bibitem{krishnan2014activity}
N.~C. Krishnan and D.~J. Cook, ``Activity recognition on streaming sensor
  data,'' \emph{Pervasive and mobile computing}, vol.~10, pp. 138--154, 2014.

\bibitem{wang2017human}
Z.~Wang, N.~Yang, M.~Guo, and H.~Zhao, ``Human-human interactional synchrony
  analysis based on body sensor networks,'' \emph{IEEE Transactions on
  Affective Computing}, vol.~10, no.~3, pp. 407--416, 2017.

\bibitem{kim2009human}
E.~Kim, S.~Helal, and D.~Cook, ``Human activity recognition and pattern
  discovery,'' \emph{IEEE pervasive computing}, vol.~9, no.~1, pp. 48--53,
  2009.

\bibitem{rashidi2010discovering}
P.~Rashidi, D.~J. Cook, L.~B. Holder, and M.~Schmitter-Edgecombe, ``Discovering
  activities to recognize and track in a smart environment,'' \emph{IEEE
  transactions on knowledge and data engineering}, vol.~23, no.~4, pp.
  527--539, 2010.

\bibitem{acampora2013survey}
G.~Acampora, D.~J. Cook, P.~Rashidi, and A.~V. Vasilakos, ``A survey on ambient
  intelligence in healthcare,'' \emph{Proceedings of the IEEE}, vol. 101,
  no.~12, pp. 2470--2494, 2013.

\bibitem{noor2017adaptive}
M.~H.~M. Noor, Z.~Salcic, I.~Kevin, and K.~Wang, ``Adaptive sliding window
  segmentation for physical activity recognition using a single tri-axial
  accelerometer,'' \emph{Pervasive and Mobile Computing}, vol.~38, pp. 41--59,
  2017.

\bibitem{okeyo2014dynamic}
G.~Okeyo, L.~Chen, H.~Wang, and R.~Sterritt, ``Dynamic sensor data segmentation
  for real-time knowledge-driven activity recognition,'' \emph{Pervasive and
  Mobile Computing}, vol.~10, pp. 155--172, 2014.

\bibitem{wang2021sensor}
J.~Wang, T.~Zhu, J.~Gan, H.~Ning, and Y.~Wan, ``Sensor data augmentation with
  resampling for contrastive learning in human activity recognition,''
  \emph{arXiv preprint arXiv:2109.02054}, 2021.

\bibitem{wang2022negative}
J.~Wang, T.~Zhu, L.~Chen, H.~Ning, and Y.~Wan, ``Negative selection by
  clustering for contrastive learning in human activity recognition,''
  \emph{arXiv preprint arXiv:2203.12230}, 2022.

\bibitem{ronao2016human}
C.~A. Ronao and S.-B. Cho, ``Human activity recognition with smartphone sensors
  using deep learning neural networks,'' \emph{Expert systems with
  applications}, vol.~59, pp. 235--244, 2016.

\bibitem{hochreiter1997long}
S.~Hochreiter and J.~Schmidhuber, ``Long short-term memory,'' \emph{Neural
  computation}, vol.~9, no.~8, pp. 1735--1780, 1997.

\bibitem{ordonez2016deep}
F.~J. Ord{\'o}{\~n}ez and D.~Roggen, ``Deep convolutional and lstm recurrent
  neural networks for multimodal wearable activity recognition,''
  \emph{Sensors}, vol.~16, no.~1, p. 115, 2016.

\bibitem{zhang2020sensors}
W.~Zhang, T.~Zhu, C.~Yang, J.~Xiao, and H.~Ning, ``Sensors-based human activity
  recognition with convolutional neural network and attention mechanism,'' in
  \emph{2020 IEEE 11th International Conference on Software Engineering and
  Service Science (ICSESS)}.\hskip 1em plus 0.5em minus 0.4em\relax IEEE, 2020,
  pp. 158--162.

\bibitem{qian2021weakly}
H.~Qian, S.~J. Pan, and C.~Miao, ``Weakly-supervised sensor-based activity
  segmentation and recognition via learning from distributions,''
  \emph{Artificial Intelligence}, vol. 292, p. 103429, 2021.

\bibitem{liu2016ssd}
W.~Liu, D.~Anguelov, D.~Erhan, C.~Szegedy, S.~Reed, C.-Y. Fu, and A.~C. Berg,
  ``Ssd: Single shot multibox detector,'' in \emph{European conference on
  computer vision}.\hskip 1em plus 0.5em minus 0.4em\relax Springer, 2016, pp.
  21--37.

\bibitem{girshick2014rich}
R.~Girshick, J.~Donahue, T.~Darrell, and J.~Malik, ``Rich feature hierarchies
  for accurate object detection and semantic segmentation,'' in
  \emph{Proceedings of the IEEE conference on computer vision and pattern
  recognition}, 2014, pp. 580--587.

\bibitem{aminikhanghahi2018real}
S.~Aminikhanghahi, T.~Wang, and D.~J. Cook, ``Real-time change point detection
  with application to smart home time series data,'' \emph{IEEE Transactions on
  Knowledge and Data Engineering}, vol.~31, no.~5, pp. 1010--1023, 2018.

\bibitem{aminikhanghahi2017survey}
S.~Aminikhanghahi and D.~J. Cook, ``A survey of methods for time series change
  point detection,'' \emph{Knowledge and information systems}, vol.~51, no.~2,
  pp. 339--367, 2017.

\bibitem{guedon2013exploring}
Y.~Gu{\'e}don, ``Exploring the latent segmentation space for the assessment of
  multiple change-point models,'' \emph{Computational Statistics}, vol.~28,
  no.~6, pp. 2641--2678, 2013.

\bibitem{chen2021deep}
K.~Chen, D.~Zhang, L.~Yao, B.~Guo, Z.~Yu, and Y.~Liu, ``Deep learning for
  sensor-based human activity recognition: Overview, challenges, and
  opportunities,'' \emph{ACM Computing Surveys (CSUR)}, vol.~54, no.~4, pp.
  1--40, 2021.

\bibitem{wang2012hierarchical}
L.~Wang, T.~Gu, X.~Tao, and J.~Lu, ``A hierarchical approach to real-time
  activity recognition in body sensor networks,'' \emph{Pervasive and Mobile
  Computing}, vol.~8, no.~1, pp. 115--130, 2012.

\bibitem{banos2014window}
O.~Banos, J.-M. Galvez, M.~Damas, H.~Pomares, and I.~Rojas, ``Window size
  impact in human activity recognition,'' \emph{Sensors}, vol.~14, no.~4, pp.
  6474--6499, 2014.

\bibitem{wan2015dynamic}
J.~Wan, M.~J. O’grady, and G.~M. O’Hare, ``Dynamic sensor event
  segmentation for real-time activity recognition in a smart home context,''
  \emph{Personal and Ubiquitous Computing}, vol.~19, no.~2, pp. 287--301, 2015.

\bibitem{li2018specialized}
H.~Li, G.~D. Abowd, and T.~Pl{\"o}tz, ``On specialized window lengths and
  detector based human activity recognition,'' in \emph{Proceedings of the 2018
  ACM international symposium on wearable computers}, 2018, pp. 68--71.

\bibitem{santos2015trajectory}
L.~Santos, K.~Khoshhal, and J.~Dias, ``Trajectory-based human action
  segmentation,'' \emph{Pattern Recognition}, vol.~48, no.~2, pp. 568--579,
  2015.

\bibitem{ma2020adaptive}
C.~Ma, W.~Li, J.~Cao, J.~Du, Q.~Li, and R.~Gravina, ``Adaptive sliding window
  based activity recognition for assisted livings,'' \emph{Information Fusion},
  vol.~53, pp. 55--65, 2020.

\bibitem{ye2015kcar}
J.~Ye, G.~Stevenson, and S.~Dobson, ``Kcar: A knowledge-driven approach for
  concurrent activity recognition,'' \emph{Pervasive and Mobile Computing},
  vol.~19, pp. 47--70, 2015.

\bibitem{triboan2017semantic}
D.~Triboan, L.~Chen, F.~Chen, and Z.~Wang, ``Semantic segmentation of real-time
  sensor data stream for complex activity recognition,'' \emph{Personal and
  Ubiquitous Computing}, vol.~21, no.~3, pp. 411--425, 2017.

\bibitem{triboan2019semantics}
------, ``A semantics-based approach to sensor data segmentation in real-time
  activity recognition,'' \emph{Future Generation Computer Systems}, vol.~93,
  pp. 224--236, 2019.

\bibitem{bianchi2019iot}
V.~Bianchi, M.~Bassoli, G.~Lombardo, P.~Fornacciari, M.~Mordonini, and
  I.~De~Munari, ``Iot wearable sensor and deep learning: An integrated approach
  for personalized human activity recognition in a smart home environment,''
  \emph{IEEE Internet of Things Journal}, vol.~6, no.~5, pp. 8553--8562, 2019.

\bibitem{gil2020improving}
M.~Gil-Mart{\'\i}n, R.~San-Segundo, F.~Fernandez-Martinez, and
  J.~Ferreiros-L{\'o}pez, ``Improving physical activity recognition using a new
  deep learning architecture and post-processing techniques,''
  \emph{Engineering Applications of Artificial Intelligence}, vol.~92, p.
  103679, 2020.

\bibitem{pham2020senscapsnet}
C.~Pham, S.~Nguyen-Thai, H.~Tran-Quang, S.~Tran, H.~Vu, T.-H. Tran, and T.-L.
  Le, ``Senscapsnet: deep neural network for non-obtrusive sensing based human
  activity recognition,'' \emph{IEEE Access}, vol.~8, pp. 86\,934--86\,946,
  2020.

\bibitem{cruciani2020feature}
F.~Cruciani, A.~Vafeiadis, C.~Nugent, I.~Cleland, P.~McCullagh, K.~Votis,
  D.~Giakoumis, D.~Tzovaras, L.~Chen, and R.~Hamzaoui, ``Feature learning for
  human activity recognition using convolutional neural networks,'' \emph{CCF
  Transactions on Pervasive Computing and Interaction}, vol.~2, no.~1, pp.
  18--32, 2020.

\bibitem{huang2021shallow}
W.~Huang, L.~Zhang, W.~Gao, F.~Min, and J.~He, ``Shallow convolutional neural
  networks for human activity recognition using wearable sensors,'' \emph{IEEE
  Transactions on Instrumentation and Measurement}, vol.~70, pp. 1--11, 2021.

\bibitem{xu2019innohar}
C.~Xu, D.~Chai, J.~He, X.~Zhang, and S.~Duan, ``Innohar: A deep neural network
  for complex human activity recognition,'' \emph{Ieee Access}, vol.~7, pp.
  9893--9902, 2019.

\bibitem{xia2020lstm}
K.~Xia, J.~Huang, and H.~Wang, ``Lstm-cnn architecture for human activity
  recognition,'' \emph{IEEE Access}, vol.~8, pp. 56\,855--56\,866, 2020.

\bibitem{tong2022novel}
L.~Tong, H.~Ma, Q.~Lin, J.~He, and L.~Peng, ``A novel deep learning bi-gru-i
  model for real-time human activity recognition using inertial sensors,''
  \emph{IEEE Sensors Journal}, 2022.

\bibitem{tang2022triple}
Y.~Tang, L.~Zhang, Q.~Teng, F.~Min, and A.~Song, ``Triple cross-domain
  attention on human activity recognition using wearable sensors,'' \emph{IEEE
  Transactions on Emerging Topics in Computational Intelligence}, 2022.

\bibitem{Khan2021}
Z.~N. Khan and J.~Ahmad, ``Attention induced multi-head convolutional neural
  network for human activity recognition,'' \emph{Applied Soft Computing}, vol.
  110, 10 2021.

\bibitem{gao2021deep}
W.~Gao, L.~Zhang, W.~Huang, F.~Min, J.~He, and A.~Song, ``Deep neural networks
  for sensor-based human activity recognition using selective kernel
  convolution,'' \emph{IEEE Transactions on Instrumentation and Measurement},
  vol.~70, pp. 1--13, 2021.

\bibitem{stiefmeier2007fusion}
T.~Stiefmeier, D.~Roggen, and G.~Troster, ``Fusion of string-matched templates
  for continuous activity recognition,'' in \emph{2007 11th IEEE International
  Symposium on Wearable Computers}.\hskip 1em plus 0.5em minus 0.4em\relax
  IEEE, 2007, pp. 41--44.

\bibitem{forster2009unsupervised}
K.~Forster, D.~Roggen, and G.~Troster, ``Unsupervised classifier
  self-calibration through repeated context occurences: Is there robustness
  against sensor displacement to gain?'' in \emph{2009 international symposium
  on wearable computers}.\hskip 1em plus 0.5em minus 0.4em\relax IEEE, 2009,
  pp. 77--84.

\bibitem{shoaib2014fusion}
M.~Shoaib, S.~Bosch, O.~D. Incel, H.~Scholten, and P.~J. Havinga, ``Fusion of
  smartphone motion sensors for physical activity recognition,''
  \emph{Sensors}, vol.~14, no.~6, pp. 10\,146--10\,176, 2014.

\bibitem{kwapisz2011activity}
J.~R. Kwapisz, G.~M. Weiss, and S.~A. Moore, ``Activity recognition using cell
  phone accelerometers,'' \emph{ACM SigKDD Explorations Newsletter}, vol.~12,
  no.~2, pp. 74--82, 2011.

\bibitem{anguita2013public}
D.~Anguita, A.~Ghio, L.~Oneto, X.~Parra~Perez, and J.~L. Reyes~Ortiz, ``A
  public domain dataset for human activity recognition using smartphones,'' in
  \emph{Proceedings of the 21th international European symposium on artificial
  neural networks, computational intelligence and machine learning}, 2013, pp.
  437--442.

\bibitem{reiss2012introducing}
A.~Reiss and D.~Stricker, ``Introducing a new benchmarked dataset for activity
  monitoring,'' in \emph{2012 16th international symposium on wearable
  computers}.\hskip 1em plus 0.5em minus 0.4em\relax IEEE, 2012, pp. 108--109.

\bibitem{micucci2017unimib}
D.~Micucci, M.~Mobilio, and P.~Napoletano, ``Unimib shar: A dataset for human
  activity recognition using acceleration data from smartphones,''
  \emph{Applied Sciences}, vol.~7, no.~10, p. 1101, 2017.

\bibitem{aminikhanghahi2019enhancing}
S.~Aminikhanghahi and D.~J. Cook, ``Enhancing activity recognition using
  cpd-based activity segmentation,'' \emph{Pervasive and Mobile Computing},
  vol.~53, pp. 75--89, 2019.

\bibitem{keogh2001online}
E.~Keogh, S.~Chu, D.~Hart, and M.~Pazzani, ``An online algorithm for segmenting
  time series,'' in \emph{Proceedings 2001 IEEE international conference on
  data mining}.\hskip 1em plus 0.5em minus 0.4em\relax IEEE, 2001, pp.
  289--296.

\bibitem{fryzlewicz2014wild}
P.~Fryzlewicz, ``Wild binary segmentation for multiple change-point
  detection,'' \emph{The Annals of Statistics}, vol.~42, no.~6, pp. 2243--2281,
  2014.

\bibitem{abedin2021attend}
A.~Abedin, M.~Ehsanpour, Q.~Shi, H.~Rezatofighi, and D.~C. Ranasinghe, ``Attend
  and discriminate: Beyond the state-of-the-art for human activity recognition
  using wearable sensors,'' \emph{Proceedings of the ACM on Interactive,
  Mobile, Wearable and Ubiquitous Technologies}, vol.~5, no.~1, pp. 1--22,
  2021.

\bibitem{varamin2018deep}
A.~A. Varamin, E.~Abbasnejad, Q.~Shi, D.~C. Ranasinghe, and H.~Rezatofighi,
  ``Deep auto-set: A deep auto-encoder-set network for activity recognition
  using wearables,'' in \emph{Proceedings of the 15th EAI International
  Conference on Mobile and Ubiquitous Systems: Computing, Networking and
  Services}, 2018, pp. 246--253.

\bibitem{wan2020deep}
S.~Wan, L.~Qi, X.~Xu, C.~Tong, and Z.~Gu, ``Deep learning models for real-time
  human activity recognition with smartphones,'' \emph{Mobile Networks and
  Applications}, vol.~25, no.~2, pp. 743--755, 2020.

\bibitem{guan2017ensembles}
Y.~Guan and T.~Pl{\"o}tz, ``Ensembles of deep lstm learners for activity
  recognition using wearables,'' \emph{Proceedings of the ACM on Interactive,
  Mobile, Wearable and Ubiquitous Technologies}, vol.~1, no.~2, pp. 1--28,
  2017.

\bibitem{yao2018efficient}
R.~Yao, G.~Lin, Q.~Shi, and D.~C. Ranasinghe, ``Efficient dense labelling of
  human activity sequences from wearables using fully convolutional networks,''
  \emph{Pattern Recognition}, vol.~78, pp. 252--266, 2018.

\bibitem{teng2020layer}
Q.~Teng, K.~Wang, L.~Zhang, and J.~He, ``The layer-wise training convolutional
  neural networks using local loss for sensor-based human activity
  recognition,'' \emph{IEEE Sensors Journal}, vol.~20, no.~13, pp. 7265--7274,
  2020.

\end{thebibliography}
\begin{IEEEbiography}[{\includegraphics[width=1in,height=1.25in, clip,keepaspectratio]{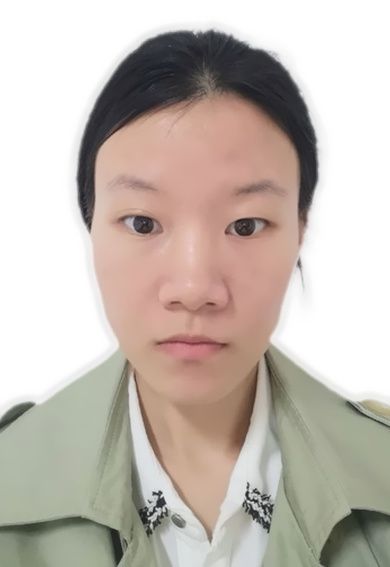}}]{Furong Duan}
	received his B.E. degree from
	Hengyang Normal University in 2019. Her is currently
	a M.S. student in the School of Computer Science,
	University of South China. His research interests include intelligent perception and pattern recognition.
	\vspace{-15 mm} 
\end{IEEEbiography}
\begin{IEEEbiography}[{\includegraphics[width=1in,height=1.25in,
		clip,keepaspectratio]{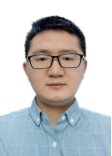}}]{Tao Zhu}
	 received his B.E. degree from Central South University, Changsha, China, and Ph.D. from University of Science and Technology of China, Hefei, China, in 2009 and 2015 respectively. He is currently an associate professor at University of South China, Hengyang, China. He is the principal investigator of several projects funded by the National Natural Science Foundation of China and Science Foundation of Hunan Province etc. He is now the Chair of IEEE CIS Smart World Technical Committee Task Force on "User-Centred Smart Systems".His research interests include IoT, pervasive computing, assisted living and evolutionary computation.
	\vspace{-10 mm}
\end{IEEEbiography}
\begin{IEEEbiography}[{\includegraphics[width=1in,height=1.25in,
		clip,keepaspectratio]{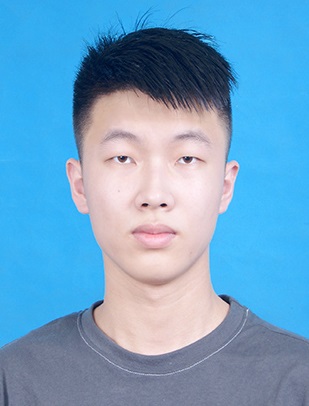}}]{Jinqiang Wang}
received his B.E. degree from
Henan Normal University in 2020. He is currently
a M.S. student in the School of Computer Science,
University of South China. His research interests include intelligent perception and pattern recognition.
	
\end{IEEEbiography}

\end{document}